\documentclass[lettersize,journal]{IEEEtran}
\usepackage{amsmath,amsfonts}
\usepackage{algorithmic}
\usepackage{algorithm}
\usepackage{array}
\usepackage[caption=false,font=normalsize,labelfont=sf,textfont=sf]{subfig}
\usepackage{textcomp}
\usepackage{stfloats}
\usepackage{url}
\usepackage{verbatim}
\usepackage{graphicx}
\usepackage{cite}
\usepackage{graphicx}
\usepackage{amsmath}
\usepackage{amssymb}
\usepackage{booktabs}
\usepackage{xcolor}
\usepackage{bm}
\usepackage{pifont}
\usepackage{enumitem}
\usepackage{soul}
\usepackage{makecell}
\usepackage{tabularx}
\usepackage{multirow}

\usepackage{siunitx}
\usepackage{balance}

\usepackage[pagebackref,breaklinks,colorlinks]{hyperref}

\def \ie {\emph{i.e.},}
\def \eg {\emph{e.g.},}
\def \etal {\emph{et al.}}

\newcolumntype{Y}{>{\centering\arraybackslash}X}

\newcommand{\tit}[1]{\smallbreak\noindent\textbf{#1.}}

\newcommand{\numfonts}{\num{10400} }
\newcommand{\numwords}{\num{10400} }

\usepackage[capitalize]{cleveref}
\crefname{section}{Sec.}{Secs.}
\Crefname{section}{Section}{Sections}
\Crefname{table}{Table}{Tables}
\crefname{table}{Tab.}{Tabs.}

\begin{document}

\title{VATr++: Choose Your Words Wisely for \\Handwritten Text Generation}

\author{Bram~Vanherle,
        Vittorio~Pippi,
        Silvia~Cascianelli,
        Nick~Michiels,\\
        Frank~Van~Reeth,
        and~Rita~Cucchiara
\thanks{B. Vanherle, N. Michiels, F. Van Reeth are with Hasselt University, Expertise center for Digial Media -- tUL -- Flanders Make, Diepenbeek, Belgium. E-mail: \{bram.vanherle, nick.michiels, frank.vanreeth\}@uhasselt.be. 
V. Pippi, S. Cascianelli, and R. Cucchiara are with the Department of Engineering ``Enzo Ferrari'', University of Modena and Reggio Emilia, Modena, Italy. E-mail: \{vittorio.pippi, silvia.cascianelli,  rita.cucchiara\}@unimore.it. 
}
\thanks{This work has been carried out during Bram Vanherle internship at the University of Modena and Reggio Emilia}
}

\maketitle

\begin{abstract}
Styled Handwritten Text Generation (HTG) has received significant attention in recent years, propelled by the success of learning-based solutions employing GANs, Transformers, and, preliminarily, Diffusion Models. Despite this surge in interest, there remains a critical yet understudied aspect – the impact of the input, both visual and textual, on the HTG model training and its subsequent influence on performance. This study delves deeper into a cutting-edge Styled-HTG approach, proposing strategies for input preparation and training regularization that allow the model to achieve better performance and generalize better. These aspects are validated through extensive analysis on several different settings and datasets.
Moreover, in this work, we go beyond performance optimization and address a significant hurdle in HTG research – the lack of a standardized evaluation protocol. In particular, we propose a standardization of the evaluation protocol for HTG and conduct a comprehensive benchmarking of existing approaches. By doing so, we aim to establish a foundation for fair and meaningful comparisons between HTG strategies, fostering progress in the field.

\end{abstract}


\section{Introduction}
\label{sec:introduction}

\IEEEPARstart{S}{tyled} handwritten text generation (HTG) is an image generation task in which arbitrary strings are rendered on images mimicking a specific person’s handwriting~\cite{bhunia2021handwriting, fogel2020scrabblegan, kang2020ganwriting}. Given a small number of handwriting samples, any text can be generated in that person’s calligraphic style. Therefore, solving this task can be useful for generating large amounts of diverse, high-quality training data for other handwriting-related tasks, such as Handwritten Text Recognition (HTR)~\cite{bhunia2021metahtr, bhunia2021text, zhang2019sequence, bhunia2019handwriting, kang2021content, kang2020distilling}. Additionally, this technique can help physically impaired people create handwritten notes in their own handwriting. Moreover, the internal representation of a person’s handwriting that is obtained as a by-product can be used for other downstream tasks such as writer identification and signature verification~\cite{pippi2023evaluating}.

\begin{figure}[h]
    \centering
    \includegraphics[width=\columnwidth]{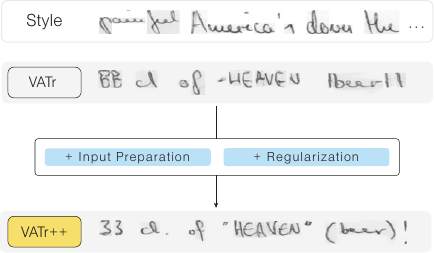} 
    \caption{We extend our previous State-of-the-Art Styled Handwritten Text Generation model \emph{VATr}~\cite{pippi2023handwritten} by training it on a modified dataset and using smart augmentations of the training signals. This enables the network to generate rare characters more faithfully and generalize to new styles.}
    \label{fig:first_page}
    \vspace{-.5em}
\end{figure}

Handwritten text generation can be approached either by considering the handwritten word as a trajectory of subsequent pen positions or as a static image. The former strategy is called \emph{Online HTG}~\cite{graves2013generating, aksan2018deepwriting, aksan2018stcn, ji2019generative, kotani2020generating}. Models trained using this paradigm can generate clean-looking text. However, it is very costly to collect the data needed to train them since the entire pen trajectory needs to be captured to represent the textual content. Moreover, to represent the writer’s style, all the parameters contributing to the way the handwriting strokes look (\eg~thickness, color, texture) also need to be captured on a sequential basis. This makes it unfeasible to use this method on historical documents, an area in which HTG can be leveraged to generate synthetic data to combat data scarcity~\cite{vogtlin2021synthetic, cascianelli2021learning}. A more popular alternative is the \emph{Offline} approach to \emph{HTG}~\cite{wang2005combining, lin2007style, thomas2009synthetic, haines2016my, alonso2019adversarial, fogel2020scrabblegan, davis2020text, kang2020ganwriting, mattick2021smartpatch, gan2021higan, gan2022higan+, bhunia2021handwriting, krishnan2021textstylebrush, luo2022slogan}.  This technique does not require costly trajectory data for training. Instead, handwriting is captured by static images that implicitly encode all the necessary style information. Consequently, the styled text is generated as a whole image. In light of these considerations, we follow the offline paradigm.

The datasets adopted for training HTG models gather a relatively large number of images containing words written by different writers. Some examples are IAM~\cite{marti2002iam}, RIMES~\cite{augustin2006rimes}, and CVL~\cite{kleber2013cvl}, with IAM being the most used one. For each writer, a few samples (usually from one to 15) are given to the model as reference style examples. A dedicated component of the HTG model extracts style features from those samples. Note that the style images are usually resized to a fixed height for ease of implementation. A different component of the HTG model is in charge of encoding the desired textual content that will appear in the generated image. The common approach is to consider a finite charset and represent each character as a one-hot vector. During training, the HTG model is asked to generate words picked from a defined corpus, which are fed to the text encoder as sequences of one-hot vectors. The resulting textual content representation is then combined with the style features and fed to another model component that outputs the styled text image. For training HTG models, most of the State-of-the-Art approaches (apart from a few preliminary works exploiting Diffusion Models~\cite{zhu2023conditional, nikolaidou2023wordstylist}) follow the adversarial learning paradigm. This means that the HTG model is the generator, and another dedicated network is used as a discriminator to distinguish the writer’s real images from the model’s generated ones. Commonly, an auxiliary HTR network is used to guide the network to generate readable text, and a writer classification model to ensure that the network maintains the author’s style in the generated image.

Whereas current HTG systems can generate realistic-looking texts, they fail at rendering long-tail and out-of-charset characters. Long-tail characters appear only rarely in the training set and are thus not often seen by the style encoder, discriminator, and HTR network. 
To face this issue, in our work~\cite{pippi2023handwritten}, we introduced the Visual Archetype Transformer (VATr). Previous HTG systems represent the text to be rendered by the model as a sequence of characters defined by a one-hot encoding. This approach fails to exploit the geometric similarities between characters and limits the charset to the ones seen during training. On the other hand, VATr represents the target text as a sequence of Visual Archetypes, 16${\times}$16 binary images of the characters rendered in the GNU Unifont. This way, prior knowledge of geometric aspects of common characters can be used to exploit similarities when generating rare and unseen characters. The architecture of VATr is a hybrid Convolutional-Transformer encoder-decoder trained by following an adversarial paradigm~\cite{goodfellow2014generative, mirza2014conditional}. The encoder uses a Convolutional Neural Network (CNN) pre-trained on a massive synthetic dataset to process each style sample individually. The large-scale pre-training helps the network generalize better to unseen styles and focus more on calligraphic aspects. Moreover, in the Transformer encoder, the self-attention mechanism is leveraged to combine the samples into a vector representing the author’s style. Then, in the Transformer decoder, the style vector is attended by the query words represented by the Visual Archetypes. The cross-attention mechanism helps capture global and local relations between content and style. Finally, a convolutional decoder generates the styled word image. 

Although improving the long-tail character generation, VATr still struggles to generate very rare characters faithfully. We argue that this weakness stems from shortcomings in the input data fed to the model during training. Therefore, in this extension to our work on VATr, we introduce several improvements to the training procedure that enhance rare character generation and favor generalization. The resulting solution is referred to as \emph{VATr++} in the following (\cref{fig:first_page}).

Through extensive experimental validation, both qualitative and quantitative, we show the benefits of the proposed modifications for generating images of handwritten text whose style is faithful to the reference, even when unseen in training, and whose content is readable, even when containing long-tail characters and unseen character combinations. The code and weights will be made available at \url{github.com/EDM-Research/VATr-pp}.
Note that, in such evaluation, we also make use of an evaluation protocol that we devise to be clearly defined and thus easily applicable to other HTG models for direct and fair comparison. The evaluation protocol comes with the necessary scripts and files to easily apply it (available at \url{github.com/aimagelab/htg_eval}), thus favoring fair and direct evaluation of HTG models.

\subsection{VATr++ Features}
The main characteristics introduced in VATr++ over our recently proposed work~\cite{pippi2023handwritten} are presented below. Although we apply them to our VATr architecture, these are general strategies to ease the training and improve the performance of few-shot styled-HTG models.
\tit{Style Input Preparation}
First, we remark that in some datasets used for HTG, small symbols such as punctuation marks are treated as separate words and scaled to the same fixed height as other words containing bigger symbols (such as characters or digits). This leads to ambiguity between different punctuation marks and inconsistency among characters of the same type. To alleviate this, we propose to make sure that the punctuation marks in the training dataset always appear attached to entire words in the images used as style samples. In this respect, we introduce a modified version of the commonly adopted IAM dataset with this characteristic and use it for training the VATr model. 
\tit{Text Input Preparation}
Second, we find that if the corpus of words that the HTG model is asked to generate in training does not feature sufficient variation, some characters will be hardly ever generated, thus contributing to the poor results on long-tail character generation. To solve this, we introduce a text augmentation method that balances the distribution of characters in the training text. 
\tit{HTR Model Regularization}
Third, we argue that if the HTR model, which is trained on real images, is not sufficiently exposed to rare characters in different styles, it will not recognize them properly at inference time, thus hindering the training of the generator and increasing the risk of mode collapse. To address this, we propose to train the HTR model with real images to which we apply content-preserving augmentations.
\tit{Discriminator Regularization}
Finally, we notice difficulty in balancing the adversarial training process. The discriminator tends to become too powerful, leading it to overfit and reject rare characters. To combat this, we propose to regularize the discriminator by randomly cropping its input. 

\subsection{HTG Evaluation Protocol}
Styled-HTG is a subject that has witnessed renewed attention only in recent years. Due to this, there is no established evaluation protocol to fairly compare approaches to each other. Evaluating a generative model involves comparing several real reference images to images created by the model. For HTG, the generated images are usually compared to those from the test set of the dataset used also in training. This comparison is done by computing a score, such as the Fréchet Inception Distance (FID)~\cite{heusel2017gans}. Although most research in this domain uses the same score to evaluate their results, the way these are computed is not standardized. The generation of styled handwriting is impacted by a number of factors, such as the word rendered and the style samples used. Furthermore, the value of the FID is impacted by the real images chosen as the reference set. In the existing styled-HTG literature, it is often unclear how styled images are generated and to which real images they are compared. This makes it difficult to compare models based on the FID scores reported in their papers. To alleviate this, in this work, we propose an evaluation protocol that defines exactly what words should be generated, which reference style images are used, and to what real images they have to be compared. We apply this protocol to several existing works to fairly assess the progress in the field.\newline

The rest of this paper is organized as follows. In~\cref{sec:related}, we overview the State of the Art of the HTG task both in terms of proposed models and evaluation strategies. In~\cref{sec:model} and~\cref{sec:training}, we describe the few-shot styled-HTG architecture and the proposed strategies to improve its performance. In~\cref{sec:evaluation}, we detail our proposed evaluation protocol for the task. In~\cref{sec:experiments}, we present the experiment setup and evaluation results. Finally,~\cref{sec:conclusion} concludes the paper.
\section{Related Work}\label{sec:related}

HTG is connected to the Font Synthesis task, which involves representing and utilizing the desired style to ensure consistent character rendering~\cite{azadi2018multi, cha2020few, park2021few, xie2021dg, lee2022arbitrary}. However, Font Synthesis methods primarily focus on generating individual characters, making them particularly relevant to HTG when dealing with ideogrammatic languages~\cite{chang2018generating, gao2019artistic, jiang2019scfont,yuan2022se}. In a broader context, whether for ideogrammatic or non-ideogrammatic languages, handwriting can be approached in two ways: as a trajectory that captures the shape of the strokes forming the characters or as a static image that captures their overall appearance. Depending on this conceptualization, online or offline HTG approaches can be employed.

\tit{Online HTG} 
Online HTG approaches rely on sequential models like LSTMs~\cite{graves2013generating}, Conditional Variational RNNs~\cite{aksan2018deepwriting}, or Stochastic Temporal CNNs~\cite{aksan2018stcn} to predict point-by-point the position of the pen, given its current state and the desired text content of the image to generate. This paradigm was first proposed in~\cite{graves2013generating}, disregarding the handwriting style. The following works~\cite{aksan2018deepwriting, aksan2018stcn, kotani2020generating} addressed this point by first decoupling style and content in reference images and then recombining these features in the generated image. Moreover, to further enhance online HTG methods, in~\cite{ji2019generative}, it has been proposed to train a discriminator alongside the prediction model for the sequence of the pen positions. This strategy is in line with the popular GAN-based approach to offline HTG. However, online HTG presents some inherent limitations. In particular, they entail handling long-range dependencies. Moreover, their training data are digital pen recordings. These are costly to collect for modern handwriting, even impossible to obtain for historical authors, and not straightforward to synthesize. Hence, we follow the offline HTG paradigm.

\tit{Offline HTG}
Early offline HTG methods~\cite{wang2005combining, lin2007style, thomas2009synthetic, haines2016my} heavily rely on manual intervention to segment text and glyphs, followed by the application of handcrafted statistics-based geometric feature extraction techniques. Then, the segmented glyphs are combined with ligatures and are rendered by adding texture and blending to the paper background. Apart from the resource-intensive human involvement required by these techniques, their main drawback lies in their inability to render glyphs and ligatures that have not been observed for a specific style. In contrast, modern learning-based methods exhibit the ability to infer styled glyphs, even in cases where they have not been directly observed in the reference style examples. Despite a few attempts to perform HTG with Diffusion Models~\cite{nikolaidou2023wordstylist,zhu2023conditional}, the most typical strategy is to leverage GANs, which can be unconditioned in the case of non-stylized HTG or conditioned on a variable number of handwriting style samples in the case of stylized HTG.

The first work on learning-based HTG~\cite{alonso2019adversarial} consists of a GAN generating fixed-sized images conditioned on the embedding of the desired text content but without conditioning on the handwriting style. The constraint on the fixed size of the output image and the fixed set of text content embeddings is a major limitation of~\cite{alonso2019adversarial}. Indeed, ideally, HTG models should be able to render any textual content, thus resulting in variable-sized images. In sight of this, the authors of~\cite{fogel2020scrabblegan} proposed a more modular approach entailing concatenating single character images, although not styled. Such modular strategy became the most commonly adopted in the following works, both non-styled and styled. Specifically, the most common approach to represent the textual content is to employ independent one-hot vectors~\cite{alonso2019adversarial, davis2020text, fogel2020scrabblegan, kang2020ganwriting, bhunia2021handwriting, gan2021higan, mattick2021smartpatch, gan2022higan+}, whose number and dimension depend on an a priori-defined charset. Despite having the advantage of being modular compared to single-vector word representations, this approach does not allow for enlarging the charset efficiently and does not allow for exploiting the geometric similarities between characters. In contrast, some works~\cite{krishnan2021textstylebrush, luo2022slogan} employ images of the desired textual content rendered in a typeface font and perform style transfer to obtain the output styled image. This way, they can preserve the geometry of characters as well as their curvature and spacing at the cost of modularity. To acquire both modularity and effectiveness in exploiting geometric similarities between characters, in~\cite{pippi2023handwritten}, we proposed to represent the textual tokens in the form of sequences of Visual Archetypes, \ie~character images rendered using the comprehensive GNU Unifont, which we also employ in this work.

When performing styled HTG, in addition to the text content, a vector representation of the desired style must also be used to condition the generation~\cite{davis2020text, kang2020ganwriting, gan2021higan, mattick2021smartpatch, gan2022higan+}. Such representations are obtained from style examples that can be entire paragraphs or lines~\cite{davis2020text}, a few words~\cite{kang2020ganwriting, bhunia2021handwriting}, or even a single word~\cite{gan2021higan, luo2022slogan, gan2022higan+}. Note that using several handwriting samples from a writer typically yields superior performance~\cite{krishnan2021textstylebrush} while remaining a cost-contained procedure. Approaches tackling styled HTG usually encode the text and the style in separate stages before using them to generate the output image. A drawback of this strategy is that it prevents modeling local, content-related style patterns. To overcome this limitation, Transformer-based solutions have been proposed~\cite{bhunia2021handwriting, pippi2023handwritten}, which allow modeling content-style entanglement via the cross-attention mechanism applied between text and style representations resulting in State-of-the-Art performance. In sight of this, in this work, we exploit the Transformer-based paradigm and build upon our proposed Transformer-based VATr model~\cite{pippi2023handwritten}.

\begin{figure*}[t]
    \centering
    \includegraphics[width=\textwidth]{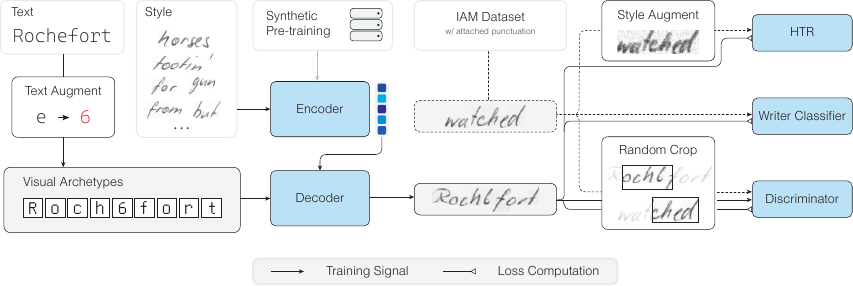}
    \caption{Overview of our extended Visual Archetypes-based Transformer for HTG (VATr++). Style samples are fed to the synthetically pre-trained style encoder, which produces a style vector for the author. The style vector is passed through the decoder along with a linear projection of the visual archetype representation of the desired text. During training, that text is augmented to increase variation. The decoder uses the style vector and text representation to generate an image. The training process is guided by a discriminator, an HTR network, and a style classification network. These networks are jointly optimized during training, utilizing real images from the IAM dataset. The input to these auxiliary networks is also augmented to prevent overfitting.}
    \vspace{-.5em}
    \label{fig:overview}
\end{figure*}

\tit{HTG Evaluation}
The vast majority of HTG models exploit multi-author word-level HTR datasets. The most commonly adopted is the IAM dataset, thanks to its size and variety. However, there is no agreement on the split to adopt, \ie~on which authors and relative images should be included in the training set and which in the test. As a result, some works adopt the standard HTR split \cite{kang2020ganwriting, bhunia2021handwriting, mattick2021smartpatch, wang2022approach, nikolaidou2023wordstylist,  pippi2023handwritten, wang2023affganwriting} (commonly known as Aachen split), while others \cite{davis2020text, fogel2020scrabblegan, gan2021higan, zdenek2021jokergan, gan2022higan+, luo2022slogan, kong2022look, zdenek2023handwritten, zhu2023conditional} consider the original split proposed with the IAM dataset, which entails a different distribution of the authors between training and test. Moreover, the text content of the generated words and the style samples considered for each author in styled-HTG are usually selected randomly, thus further hindering the fair comparison also between approaches adopting the same IAM splitting. In sight of this, in this paper, we propose to adopt a precisely defined, deterministic split of the IAM dataset to be used for HTG.

The common practice to measure the performance of HTG models is to follow Kang et al.~\etal~\cite{kang2020ganwriting} and compute the FID score on the initial crop of the generated and reference images. Other adopted scores are the Geometry Score~\cite{khrulkov2018geometry} and the Kernel Inception Distance (KID)~\cite{binkowski2018demystifying}. These scores should capture the realism of the generated images. To evaluate their readability, the Character Error Rate (CER) is also adopted. Efforts towards the definition of a task-specific score for HTG have recently brought to the vFID score~\cite{kang2021content} and the Handwriting Distance (HWD) score~\cite{pippi2023hwd}. In this work, we consider a number of evaluation scores and settings to favor the adoption of our proposed evaluation protocol and to provide a meaningful comparison between existing HTG approaches.

\section{Model Architecture}\label{sec:model}
In this work, we consider the few-shot styled variant of the offline HTG task. In this setting, we consider $P$ images containing handwritten words in the style of a certain writer of interest, $\mathrm{w}{{\in}}\mathrm{W}$. These are gathered in a set $\mathbf{X}_{\mathrm{w}}{=}\{\mathbf{x}_{\mathrm{w},i}\}_{i=0}^P$, which contains $P{=}15$ elements in this work, in accordance with~\cite{kang2020ganwriting,bhunia2021handwriting,pippi2023handwritten}. Additionally, we consider a set of $Q$ words $\mathbf{C}{=}\{\mathbf{c}_{i}\}_{i=0}^Q$ of arbitrary length $n_{i}$. The task consists of generating $Q$ images $\mathbf{Y}_{\mathrm{w}}^{\mathbf{C}}$ containing the words in $\mathbf{C}$ rendered in the style of writer $\mathrm{w}$. To tackle the task, we build upon our work~\cite{pippi2023handwritten}, where we presented an approach based on a hybrid Convolutional-Transformer architecture and a content tokens representation consisting of Visual Archetypes, dubbed VATr. We extend the approach by proposing input processing and training strategies to enhance its performance. In the following, we recall the HTG approach adopted in~\cite{pippi2023handwritten} and give the details on the proposed performance enhancement strategies.

\tit{Architecture Overview} 
We propose a Transformer encoder-decoder framework combined with a pre-trained convolutional feature extractor to manage style samples $\mathbf{X}_{\mathrm{w}}$ and with a representation of the textual content $\mathbf{C}$ consisting of character images, \ie~Visual Archetypes. In the \emph{Style Encoder} $\mathcal{E}$, the feature extractor operates on the style samples $\mathbf{X}_{\mathrm{w}}$, and its output is fed to a Transformer encoder. The encoder’s self-attention mechanism enhances these vectors by incorporating long-range dependencies and outputs a sequence of style vectors $\mathbf{S}_{\mathrm{w}}$. The \emph{Content-Guided Decoder} $\mathcal{D}$ consists of a Transformer-decoder and a convolutional decoder. In the Transformer decoder, the cross-attention is performed between the style vectors $\mathbf{S}_{\mathrm{w}}$ and the content strings $\mathbf{C}$ to be rendered, represented as a sequence of Visual Archetypes. This mechanism leads to an entangled content-style representation, enhancing local style patterns and global word appearance. The resulting representation is passed to the convolutional decoder, which generates the word images conditioned on content and style, $\mathbf{Y}_{\mathrm{w}}^{\mathbf{C}}$. An overview of the VATr++ architecture and training strategy is depicted in~\cref{fig:overview}.

\subsection{Style Encoder}
The Style Encoder $\mathcal{E}$, which is responsible for converting a small set of sample images $\mathbf{X}_{\mathrm{w}}$ into style features $\mathbf{S}_{\mathrm{w}}$, consists of a convolutional encoder and a Transformer encoder. This choice is driven by the data efficiency displayed by convolutional neural networks in extracting significant features and the suitability of the multi-head self-attention mechanism for modeling extensive dependencies within the style images. The chosen convolutional encoder backbone is ResNet18~\cite{he2016deep}, which is commonly adopted by approaches handling text images~\cite{javidi2020deep, zhu2020point, bhunia2021handwriting, manna2022swis}. In this work, we pre-train it to extract robust features from the style sample images, as detailed in~\cref{sssec:font2}. Once pre-trained, the backbone is used to extract $P$ feature maps $\mathbf{h}_{\mathrm{w},i}{{\in}}\mathbb{R}^{h{\times}w{\times}d}$ from the $P$ style images $\mathbf{x}_{\mathrm{w},i}{{\in}}\mathbf{X}_{\mathrm{w}}$. These feature maps are flattened along the spatial dimension, resulting in a $(h{\cdot}w)$-length sequence of $d$-dimensional vectors. Notably, while $h$ and $w$ vary based on the input image shape, the embedding size $d$ remains fixed and is set to 512 in this study. The elements within this sequence correspond to adjacent regions within the original images, aligning with the receptive field of the convolutional backbone. The sequences of flattened feature maps of each style image are concatenated to form the sequence $\mathbf{H}_{\mathrm{w}}{{\in}}\mathbb{R}^{N{\times}d}$, where $N{=}h{\cdot}w{\cdot}P$. This sequence is then fed into the initial layer of the multi-layer, multi-headed self-attention encoder consisting of $L{=}3$ layers, each with $J{=}8$ attention heads and a multi-layer perception. The final layer outputs the sequence of style features for writer $\mathrm{w}$, $\mathbf{H}^L{=}\mathbf{S}_{\mathrm{w}} {{\in}} \mathbb{R}^{N{\times}d}$, which is subsequently fed to the Transformer decoder within $\mathcal{D}$.

\subsubsection{Synthetic Pre-training}\label{sssec:font2}
Encouraged by the benefits of large-scale pre-training in other learning tasks and the results obtained in our preliminary work \cite{pippi2023handwritten, pippi2023evaluating}, we resort to this strategy by employing a large synthetic dataset that covers glyph shapes and ink and background textures. The dataset contains over 100 million samples and has been obtained by rendering \numwords English words in \numfonts calligraphic fonts on different paper-like backgrounds, with added realism through random geometric and color transformations. The convolutional backbone of the Style Encoder is trained on this dataset by minimizing a Cross-Entropy Loss to recognize the calligraphic font. This way, the network is encouraged to disregard general image characteristics such as text, background, and ink type, and thus, it can extract robust style features.

\subsubsection{Style Input Preparation}
\begin{figure}
    \centering
    \includegraphics[width=\columnwidth]{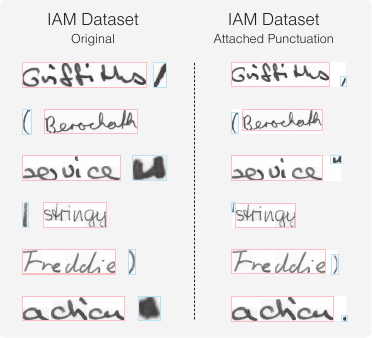}
    \caption{We modify the IAM dataset by attaching single punctuation marks (blue) to their closest words in the line-level IAM (red). This style input preparation strategy helps prevent ambiguity and inconsistency.}
    \label{fig:iam_attached}
    \vspace{-.5em}
\end{figure}

As mentioned above, the largest and most commonly used dataset for training HTG models is IAM~\cite{marti2002iam}. The dataset contains images of handwritten English words at the page, line, and word levels. At the word level, most punctuation marks are considered as standalone words. The pre-processing pipeline used in most of the State-of-the-Art HTG models~\cite{fogel2020scrabblegan, bhunia2021handwriting}, including VATr, entails resizing each word to a fixed height. As a result, punctuation marks lose their scale and spatial context. This leads to ambiguity, as different types of punctuation marks end up appearing very similar to each other. Moreover, this introduces inconsistency between the punctuation marks since those that have been left attached to words are smaller than their standalone counterparts. Some examples of these issues are reported in~\cref{fig:iam_problems}. More information and examples of these single punctuation marks are provided in Appendix~\ref{app:single_punct}.
\begin{figure}
    \centering
    \includegraphics[width=\columnwidth]{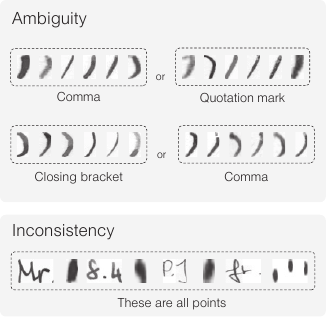}
    \caption{Punctuation marks are considered words in the IAM dataset as used for HTG. This causes ambiguity and inconsistency since both characters and punctuation marks are all scaled to the same height.}
    \label{fig:iam_problems}
    \vspace{-.5em}
\end{figure}
These factors impact the style representation of the generator, which is presented with unrealistic style samples that have to be used to capture how the writer of interest writes punctuation marks. Moreover, since most HTG models use an auxiliary HTR network to guide the generator to generate readable text, the ambiguity and inconsistency in the training dataset make it difficult for the HTR network to distinguish between certain punctuation marks. If the HTR network cannot recognize a certain character, there is no signal for the generator to generate that character properly. To combat this, we propose attaching all images containing a single punctuation mark to a neighboring word. Depending on the punctuation mark, it is attached to the previous or next word. The bounding boxes provided in the IAM dataset are used to attach the punctuation marks correctly, and the transcriptions are updated accordingly. This operation gives spatial context and relative scaling back to the punctuation marks.  \cref{fig:iam_attached} shows some examples of how the punctuation marks are attached.

\subsection{Content-Guided Decoder}
The first part of the \emph{Content-Guided Decoder} $\mathcal{D}$ is a multi-layer, multi-head decoder. The decoder has $L{=}3$ layers and $J{=}8$ heads, similar to the encoder. In this decoder, first, self-attention is performed among the content vectors representing elements within $\mathbf{C}$. Then, cross-attention is performed, treating the sequence of content vectors as queries and the style vectors $\mathbf{S}_{\mathrm{w}}$ as keys and values. This approach allows the model to learn content-style interdependencies by directing each query to focus on the style vectors that are most useful for rendering its shape. The output of the last layer for the content string $\mathbf{c}_{i}$ is a tensor $\mathbf{F}_{c_{i}}{\in}\mathbb{R}^{k_i{\times}d}$. To introduce greater variability in the generated images, we apply normal Gaussian noise to $\mathbf{F}_{c_{i}}$. Subsequently, we project it into a $(k_i{\times}\num{8192})$ matrix, which is then reshaped into a $512{\times}4{\times}4k_i$ tensor. This tensor serves as input to a convolutional decoder consisting of four residual blocks and a $\mathrm{tanh}$ activation function, which produces the styled word images $\mathbf{Y}_{\mathrm{w}}^{\mathbf{C}}$.

\subsubsection{Visual Archetypes}
In contrast to existing methods that represent content queries as embeddings derived from independent one-hot-encoded characters, our approach, as proposed in our previous work~\cite{pippi2023handwritten}, is to use a representation that captures the similarities between characters. Specifically, we derive the content queries as follows. Initially, we render the characters using the GNU Unifont font, which encompasses all Unicode characters, thus obtaining $16{\times}16$ binary images, which we then flatten and linearly project to obtain query embeddings in a $d$-dimensional space. In this way, we can leverage geometric similarities among characters, which proves beneficial in generating long-tail characters, \ie~those seldom encountered during training. Note that the HTG network can only memorize mere content-shape associations when it is supplied with independent tokens as content queries, such as the one-hot vectors. Thus, associations between characters shapes cannot be adequately built if the model is not given enough examples in training, leading to subpar generation performance for long-tail characters. Conversely, our Visual Archetypes-based input enables the network to exploit geometric characteristics and similarities between frequently occurring and long-tail characters, facilitating more faithful rendering and generation of the latter. For more insights into the Visual Archetypes, please refer to Appendix~\ref{app:archetypes}.

\subsubsection{Text Input Preparation}
\begin{figure}
    \centering
    \includegraphics[width=0.6\columnwidth]{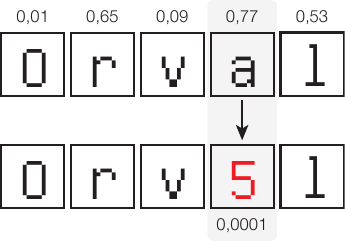}
    \caption{Text augmentation is used to increase the number of rare characters during training. Here, the word \textit{Orval} is augmented. The letter \textit{a} gets chosen to be swapped due to its high occurrence in the training corpus and is replaced by the character \textit{5}, which is rare in the dataset.}
    \label{fig:text_aug_example}
    \vspace{-.5em}
\end{figure}

State-of-the-Art HTG models, including VATr, are trained by asking the generator to generate images of certain words. The loss is computed on these images, and the model is updated accordingly. Although the words the model is asked to generate during training can be any word, in most cases, these are sampled from an unbalanced set containing very few punctuation marks, numbers, and capital letters. Due to this, the model is asked to generate certain characters only sporadically. Therefore, the model receives little feedback on the generation, resulting in difficulties when generating these characters. To train the model on a better-balanced corpus, we propose altering the words of the standard English corpus at training time and devise a specific augmentation scheme (see~\cref{fig:text_aug_example}). In our strategy, we do not change the words too much, as it would make the generated words look too different from the words in the IAM dataset, Thus facilitating the discriminator too much in distinguishing them from the real ones. To achieve this, we compute a probability set over the alphabet of the corpus:
\begin{equation*}
    P = \bigr[ \dfrac{t_\mathtt{a}}{n}, \dfrac{t_\mathtt{b}}{n}, \ldots \bigr],
\end{equation*}
with $t_i$  being the number of times character $i$ occurs and $n$ the total number of characters. This probability set is then normalized to ensure that the largest probability is $1.0$. During training, the probability with which character $c_i$ in the content string $\mathbf{c}$ is swapped is given by $\alpha P_k$, with $\alpha$ being a hyper-parameter determining the strength and $k$ the character at $c_i$. Note that setting $\alpha$ too high will make the network generate gibberish during training, while a too-low value will not increase the occurrence of rare characters enough. In light of this, we use strength \num{0.4} in our experiments. The character $c_i$ is replaced by another character, randomly sampled from the alphabet with probability $1 {-} P_j$ for character $j$. This strategy gives frequently occurring characters a high chance to be replaced by rare characters. Refer to Appendix~\ref{app:input_prep} for more information on this method.

\section{Model Training}\label{sec:training}
In a formal sense, our VATr++ model can be denoted as $\mathcal{G}_{\theta}=\mathcal{E}\circ\mathcal{D}:(\mathbf{X}_{\mathrm{w}},\textbf{C})\rightarrow\mathbf{Y}_{\mathrm{w}}^{\mathbf{C}}$. Its training is conducted concurrently with other auxiliary networks used for computing the overall loss for the generator $\mathcal{G}_{\theta}$.

The main auxiliary module is a convolutional discriminator $\mathcal{D}_{\eta}$, tasked with distinguishing the real images from those generated by $\mathcal{G}_{\theta}$. This process encourages the generator to produce images that closely resemble real ones. To enhance the performance of both $\mathcal{G}_{\theta}$ and $\mathcal{D}_{\eta}$, we adopt the adversarial paradigm utilizing the hinge adversarial loss~\cite{lim2017geometric}
\begin{equation*}
\begin{split}
    L_{adv}=& \mathbb{E} \left[\text{max}(1 - \mathcal{D}_{\eta}(\mathbf{X}_{\mathrm{w}}), 0)\right] + \\
    & \mathbb{E} \left[\text{max}(1 + \mathcal{D}_{\eta}(\mathcal{G}_{\theta}(\mathbf{X}_{\mathrm{w}}, \textbf{C})), 0)\right].
\end{split}
\end{equation*}

Moreover, we leverage an HTR model~\cite{shi2016end}, denoted as $\mathcal{R}_{\phi}$, responsible for recognizing the text in the generated images. This ensures that the generator, $\mathcal{G}_{\theta}$, not only replicates the desired stylistic features but also reproduces the textual content accurately. The HTR model is trained by using the real images $\mathbf{X}_{\mathrm{w}}$ and their corresponding ground truth transcriptions. The loss of the HTR model, calculated on the generated images $\mathbf{Y}_{\mathrm{w}}^{\mathbf{C}}$, is backpropagated through the generator $\mathcal{G}_{\theta}$. The loss of the HTR model is defined as
\begin{equation*}
    L_{HTR}=\mathbb{E}_{\mathbf{x}}\left[ - \sum \text{log}(p(t_{\mathbf{x}} | \mathcal{R}_{\phi}(\mathbf{x})))\right],
\end{equation*}
Here, $\mathbf{x}$ can represent either a real or a generated image, and $t_{\mathbf{x}}$ is its transcription. For the real images ($\mathbf{x}{\in}\mathbf{X}_{\mathrm{w}}$), $t_{\mathbf{x}}$ is obtained from the ground truth label, while for the generated images ($\mathbf{x}{\in}\mathbf{Y}_{\mathrm{w}}^{\mathbf{C}}$), it comes from $\mathbf{C}$.

Furthermore, a convolutional classifier, denoted as $\mathcal{C}_{\psi}$, is employed to classify both real and generated images based on their calligraphic style, \ie~the style of writer $\mathrm{w}$. This compels the generator, $\mathcal{G}_{\theta}$, to reproduce the intended style faithfully. Similar to the training procedure for the $\mathcal{R}_{\phi}$ module, the classifier is trained by using the real images, and its loss on the generated images guides the generator. Specifically, the loss for this module is defined as
\begin{equation*}
    L_{class}=\mathbb{E}_{\mathbf{x}}\left[ - \sum \text{log}(p(\mathrm{w} | \mathcal{C}_{\psi}(\mathbf{x})))\right].
\end{equation*}
In this context, $\mathbf{x}$ belongs to either $\mathbf{X}_{\mathrm{w}}$ or $\mathbf{Y}_{\mathrm{w}}^{\mathbf{C}}$.

To further encourage the generation of images in the desired style, we use an additional regularization loss, termed the cycle consistency loss: 
\begin{equation*}
    L_{cycle}=\mathbb{E} \left[ \left\lVert \mathcal{E}(\mathbf{X}_{\mathrm{w}}) - \mathcal{E}(\mathbf{Y}_{\mathrm{w}}^{\mathbf{C}}) \right\rVert_1 \right].
\end{equation*}
The objective is to compel the generator to produce styled images from which the encoder $\mathcal{E}$ extracts the same style vectors as from the reference images. In other words, we aim to preserve the style features of the input images in the generated ones.

In summary, the objective function used for training our VATr++ model is obtained by combining the loss terms above with equal weights, as follows:
\begin{equation*}
    L = L_{adv} + L_{HTR} +  L_{class} + L_{cycle}.
\end{equation*}

\subsection{Auxiliary Networks Regularization}
Due to their low occurrence, the auxiliary networks can show a negative bias against long-tail characters. This hinders the training process for generating these characters. To alleviate this problem, we introduce two augmentation techniques and regularize the most affected auxiliary networks.

\subsubsection{HTR Model Regularization}
\begin{figure}
    \centering
    \includegraphics[width=\columnwidth]{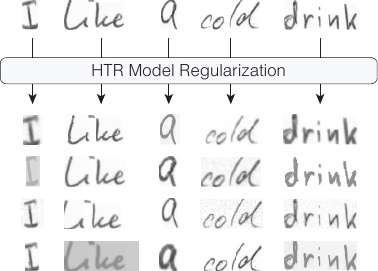}
    \caption{Inputs to the HTR network are augmented to regularize the network so that it would not overfit on specific styles.}
    \label{fig:htr_augment}
    \vspace{-.5em}
\end{figure}

During the generator training, the HTR network is trained on real images from the IAM dataset. Since some characters appear rarely in the training set, the HTR network sees only a few examples of them, possibly only from a few authors. This leads the network to overfit on these author's styles for those characters. Thus, when the HTG network tries to generate those characters in other styles, the HTR network is unable to read them. As a result, the generator is hindered from learning to produce readable long-tail characters. To solve this, we propose a regularization strategy for the HTR models, which entails augmenting the training images before feeding them to the HTR network. We select several image augmentations that change the author's style without impacting the readability of the word in the image. The augmentations used are rotation, translation, resize, elastic transform, color jitter, Gaussian blur, Gaussian noise, erosion, and dilation. All augmentations are used with small factors to preserve readability. For each training image, we combine and apply three of the mentioned augmentations with probability $0.5$. Some examples are shown in~\cref{fig:htr_augment}. Please refer to Appendix~\ref{app:htr_reg} for further details.

\begin{figure}
    \centering
    \includegraphics[width=\columnwidth]{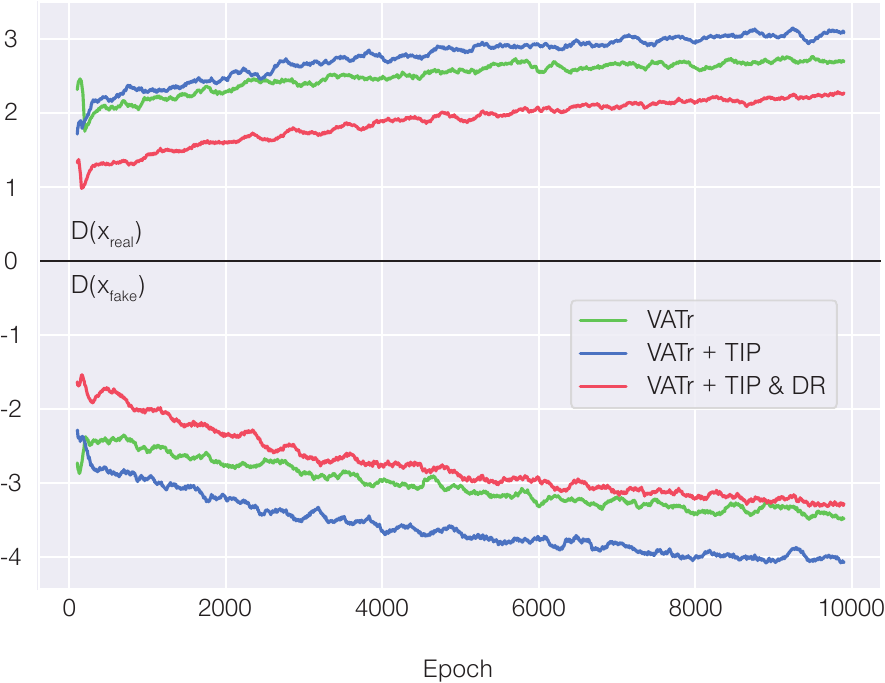}
    \caption{The averaged and smoothed responses of the discriminator network on fake and real images show that adding Text Input Preparation (TIP) makes it easier for the network to distinguish the generated images from the real ones. This can be eased using Discriminator Regularization (DR).}
    \label{fig:d_spread}
    \vspace{-.5em}
\end{figure}

\subsubsection{Discriminator Regularization}
\begin{figure}
    \centering
    \includegraphics[width=\columnwidth]{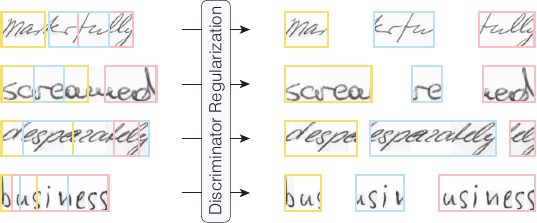}
    \caption{Some examples of the discriminator augmentation applied to some words of the IAM dataset.}
    \label{fig:discriminator_augmentation}
    \vspace{-.5em}
\end{figure}

Applying our proposed text input preparation procedure on the strings to be generated increases the quality of the generated rare characters. Nonetheless, during training, the generated words can look different from standard English ones. This makes it easier for the discriminator to distinguish between real and fake words, as they contain rare characters more often. Moreover, this can hurt the training progress, as there is smaller feedback for the generator (see~\cref{fig:d_spread}). To solve this issue and further improve the generation performance, we propose augmenting the images before feeding them to the discriminator. In particular, we apply cropping by maintaining the height but obtaining a random width between 64 and 128 pixels, sampled uniformly. The crop is taken randomly from the parts of the image containing ink to avoid obtaining completely white images (some examples are given in~\cref{fig:discriminator_augmentation}). Showing the discriminator only a local part of the word makes it more difficult for the network to distinguish between the augmented generated images and the real images of English words. To avoid the augmentations leaking to the generated images, we follow the method of Karras~\etal~\cite{karras2020limited}. Namely, we do not crop the majority of the images shown to the discriminator to ensure that the generator has to reproduce the real distribution to fool the discriminator. Specifically, augmentations are applied only 30\% of the time. Moreover, we apply the random crop to both the real and the generated images. Appendix~\ref{app:disc_reg} contains more insights in the discriminator bias and regularization.

\section{Evaluation Protocol}\label{sec:evaluation}
The absence of a standard evaluation protocol in HTG poses challenges in objectively comparing the performance of different approaches. Establishing a clear procedure for generation and evaluation is essential for ensuring consistent and meaningful assessments in this task. Moreover, developing a standardized evaluation protocol can enhance the reliability and transparency of assessments, allowing for more effective comparisons and improvements of HTG models. In this section, we describe our proposed evaluation protocol for assessing the performance of HTG models. Note that the procedure can be applied to any dataset; nonetheless, to make the description clearer, we consider the IAM dataset as a reference when defining the protocol since it is the most commonly adopted in HTG. The IAM dataset~\cite{marti2002iam} is a comprehensive collection of handwritten text samples obtained from 657 writers. The original split proposed by the authors of the dataset divides the images writers-wise into Train set, Validation 1, Validation 2, and Test set, making sure that all the images of the same writer are in only one of the subsets. The HTG community recently adopted a different split defined by GANWriting~\cite{kang2020ganwriting}, which entails dividing the IAM dataset into training (339 writers) and test sets (161 writers). For our evaluation protocol, we use this split. To ensure a consistent and fair evaluation process, we define a fixed set of reference style images and desired text to guide the generation. The evaluation encompasses various settings, described below.
\tit{In Vocabulary words, Seen writers (IV-S)} In this setting, the HTG model is tasked to generate words present in the training set with the style guide of writers used during training.
\tit{In Vocabulary words, Unseen writers (IV-U)} In this setting, the HTG model is tasked to generate words present in the training set with the style guide of writers from the test set.
\tit{Out Of Vocabulary words, Seen writers (OOV-S)} In this setting, the HTG model is tasked to generate words not present in the training set in the style of writers seen in training.
\tit{Out Of Vocabulary words, Unseen writers (OOV-U)} In this setting, the HTG model is tasked to generate words not present in the training set in the style of the test set writers.

For the above settings, we define 500 in-vocabulary words and 500 out-of-vocabulary words. The former are words from the training set, while the latter are words not present in the training set, sampled from an external English corpus. Note that we consider an English corpus in the case of IAM. In general, out-of-vocabulary words can be sampled from a corpus either in the same language as the dataset of interest or in a different language with the same characters.

\tit{Test} Finally, we define a setting in which the HTG model under evaluation is tasked to generate a replica of the test set. In this scenario, the images are generated in multiple iterations, in which the words to generate are those included in the test set, and the reference style comes from images of the same writers, containing words that are different from those to generate in the current iteration.

After the generation, the evaluation involves comparing the generated images of each setting with the corresponding real images of the writers used as style references. In our comparisons, we adopt three scores: FID and KID to measure the visual similarity and HWD to measure the calligraphic style similarity between the real and generated text images.
\section{Experiments}\label{sec:experiments}
In this section, we experimentally validate our proposed approach. First, we quantitatively compare our method to several state-of-the-art methods on the IAM dataset. Furthermore, we explore the ability of our model to generalize to unseen words, styles, and datasets. Moreover, we specifically investigate the performance of the model on rare character generation. Finally, we perform an ablation analysis to highlight the individual impact of the different performance enhancement strategies that are introduced in the model. Note that, in our experiments, we employ our proposed evaluation protocol for HTG.

\tit{Implementation Details}
The complete HTG model is trained on the IAM dataset using the Adam optimizer with a fixed learning rate of $2{\cdot}10^{-4}$ and batch size equal to \num{8}. For the discriminator and writer classification networks, we adopt the same architectures as the discriminator in~\cite{brock2019biggan}. As for the HTR network, we employ the same Convolutional-Recurrent architecture as the commonly-adopted CRNN~\cite{shi2016end}. We train our model for \num{10000} epochs while evaluating the FID every \num{500} epochs and select the best-performing model.

As for the pre-trained convolutional style encoder, we employ the same training strategy as presented in~\cite{pippi2023evaluating,pippi2023handwritten}.

\subsection{Comparison with the State-of-the-Art}
To get an idea of the competitiveness of our model, we compare it to several other State-of-the-Art styled-HTG works. Namely, we compare against the Transformer-based Handwriting Transformer (HWD)~\cite{bhunia2021handwriting}  and VATr~\cite{pippi2023handwritten} models and the recently proposed HiGAN~\cite{gan2021higan} and HiGAN+~\cite{gan2022higan+} models. For a fair comparison, we use the implementation and weights provided by the authors of the original papers. Moreover, for evaluating different aspects of the generation performance, we consider multiple scores: FID, HWD, KID, and CER. Finally, we employ the devised Test setting described in~\cref{sec:evaluation}.

For a more rigorous comparison, note that the considered methods have been trained on versions of the IAM dataset that differ for the preprocessing applied to the style images. In particular, HiGAN has been trained on the original IAM dataset without any scaling (we denote this version as \textbf{IAM-W}). HiGAN+, HWT, and VATr have been trained on a version of IAM in which the word images are scaled so that the average character width is 16 pixels (we call this version \textbf{IAM-W16}). Finally, VATr++ is also trained on scaled words images, but with the style input preparation strategy for which the single punctuation marks are attached to the words (we refer to this version as \textbf{IAM-WATTP}). In our experiments, we test all the models on all these variants of the IAM dataset, regardless of the version they have been trained on. For completeness, we also test on the scaled version of IAM, from which we remove all the images containing singleton punctuation marks (\textbf{IAM-WNOP}) to show their impact on the score.

\begin{figure*}[t]
    \centering
    \includegraphics[width=\textwidth]{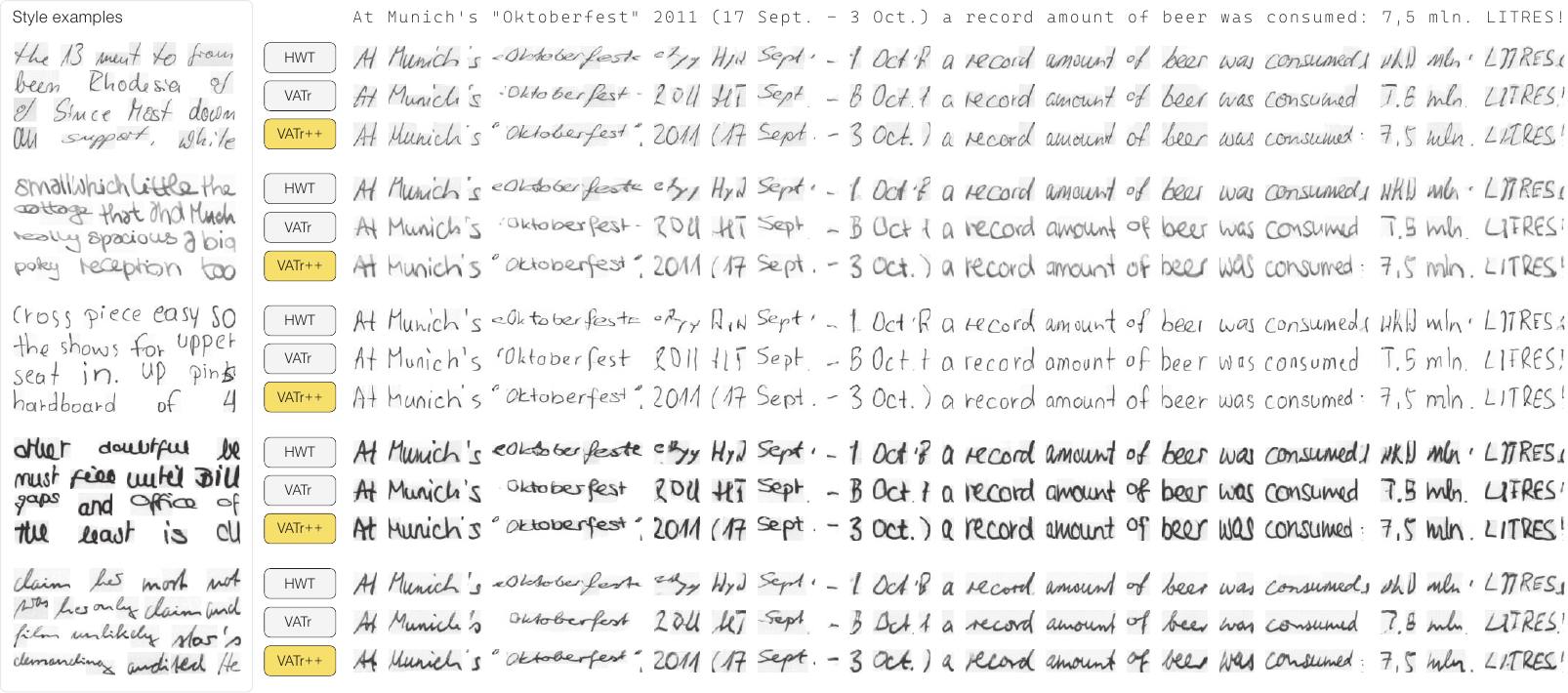}
    \caption{Qualitatives results showing the same sentence generated word-by-word with different Transformer-based HTG models in different styles.}
    \label{fig:qualitatives}
    \vspace{-.5em}
\end{figure*}

Quantitative results in terms of FID, HWD, and KID metrics are reported in~\cref{tab:fid_comp,tab:kid_comp,tab:hwd_comp}, respectively. It can be observed that for all versions of the test dataset with width scaling, VATr++ outperforms the other methods in terms of FID and KID and is best or second-best in terms of HWD. When not scaling the images, \ie~in the IAM-W setting, we notice that HiGAN is the best performer in terms of the KID and FID scores. This is expected since HiGAN has been trained on the non-scaled images. In terms of HWD, VATr is the best performer in this variant of the IAM dataset. From the tables, we also observe that there is a large difference between the FID scores on IAM-W16 and on IAM-WNOP and IAM-WATTP. This demonstrates that the presence of images with single punctuation marks has a significant impact on the FID score. The values of the HWD and KID scores follow the same trend, although less pronounced, which enforces the superior robustness of these scores compared to the FID for measuring the performance of HTG models. Additionally, some qualitative results are reported in~\cref{fig:qualitatives}, where we show that our approach can faithfully reproduce different handwriting styles and can generate a wider set of characters more accurately than the competitors. Qualitative results for these experiments are reported in Appendix~\ref{app:results}.
\begin{table}[t!]
    \footnotesize
    \centering
    \setlength{\tabcolsep}{.38em}
    \caption{Comparison between different HTG methods in terms of the FID score on the variants of the IAM dataset.}
    \label{tab:fid_comp}
    \resizebox{\linewidth}{!}{
    \begin{tabular}{l c cccc}
    \toprule
                             && \textbf{IAM-W}    & \textbf{IAM-W16}  & \textbf{IAM-WNOP} & \textbf{IAM-WATTP} \\
    \midrule    
    \textbf{HiGAN}           && \textbf{26.22}    & 19.77             & 13.99             & 13.02              \\
    \textbf{HiGAN+}          && 28.14             & 24.09             & 15.19             & 15.19              \\
    \textbf{HWT}             && 132.65            & 19.74             & 12.62             & 12.72              \\
    \textbf{VATr}            && 129.11            & 17.55             & 9.15              & 9.12               \\
    \textbf{VATr++}          && 137.61            & \textbf{16.29}    & \textbf{8.21}     & \textbf{8.27}      \\
    \bottomrule
    \end{tabular}}
\end{table}
\begin{table}[t!]
    \footnotesize
    \centering
    \setlength{\tabcolsep}{.38em}
    \caption{Comparison between different HTG methods in terms of the KID score (values are multiplied by $10^2$ for readability) on the variants of the IAM dataset.}
    \label{tab:kid_comp}
    \resizebox{\linewidth}{!}{
    \begin{tabular}{l c cccc}
    \toprule
                    && \textbf{IAM-W} & \textbf{IAM-W16} & \textbf{IAM-WNOP} & \textbf{IAM-WATTP}  \\
    \midrule
    \textbf{HiGAN}  && \textbf{~0.99} & ~0.95            & ~0.85             & ~0.74               \\
    \textbf{HiGAN+} && ~1.40          & ~1.26            & ~1.08             & ~1.08               \\
    \textbf{HWT}    && 11.55          & ~0.93            & ~0.69             & ~0.72               \\
    \textbf{VATr}   && 10.76          & ~0.59            & ~0.45             & ~0.45               \\
    \textbf{VATr++} && 11.78          & \textbf{~0.50}   & \textbf{~0.38}    & \textbf{~0.40}      \\
    \bottomrule
    \end{tabular}}
\end{table}
\begin{table}[t!]
    \footnotesize
    \centering
    \setlength{\tabcolsep}{.38em}
    \caption{Comparison between different HTG methods in terms of the HWD score on the variants of the IAM dataset.}
    \label{tab:hwd_comp}
    \resizebox{\linewidth}{!}{
    \begin{tabular}{l c cccc}
    \toprule
                    && \textbf{IAM-W}    & \textbf{IAM-W16}  & \textbf{IAM-WNOP} & \textbf{IAM-WATTP} \\
    \midrule
    \textbf{HiGAN}  && 1.04              & 1.13              & 1.21              & 1.17               \\
    \textbf{HiGAN+} && 0.72              & 0.80              & \textbf{0.61}     & 0.78               \\
    \textbf{HWT}    && 0.64              & 1.13              & 0.95              & 0.92               \\
    \textbf{VATr}   && \textbf{0.62}     & 0.83              & 0.89              & 0.86               \\
    \textbf{VATr++} && 0.63              & \textbf{0.74}     & 0.78              & \textbf{0.75}      \\
    \bottomrule  
    \end{tabular}}
\end{table}
\tit{Line-level Generation}
It is worth noting that the considered competitors and our approach have been trained at word-level. In other words, the input style samples consist of images of single words, and the text content the models are asked to generate consists of single words. As a result, these HTG models can struggle when asked to generate long text images, possibly containing multiple words as in a complete text line. Nonetheless, this ability is of high interest for practical applications. TTo assess the capability of the considered HTG models in generating text lines despite having been trained on single words, we consider a further variant of the IAM dataset, in which we take images of full text lines as given in the original dataset (we refer to this variant as \textbf{IAM-L}). Note that for the images of this variant, we apply resizing so that the characters occupy, on average, \num{16} pixels, as in the IAM-W16, IAM-WNOP, and IAM-WATTP settings. When testing on this setting, the models are asked to generate images of lines in one go by taking single words images as style samples.

\begin{figure}[t]
    \centering
    \includegraphics[width=\columnwidth]{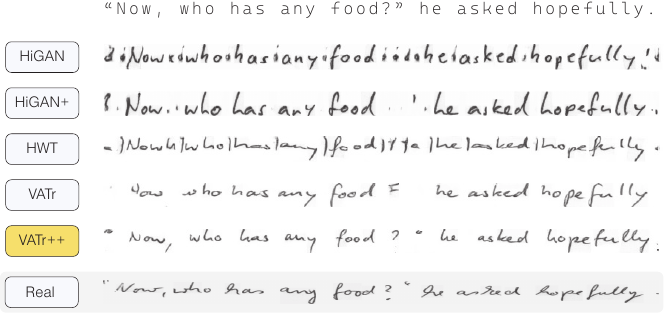}
    \caption{A line from the IAM dataset, rendered by different HTG systems. The style should be the same as the line in the last row and the text should read \textit{``Now, who has any food?'' he asked hopefully.}}
    \label{fig:line_qualitative}
    \vspace{-.5em}
\end{figure}

The results are reported in~\cref{tab:linelevel_comp}. It emerges that VATr and VATr++ are the best and second-best in most scores. Arguably, this is because the aforementioned training strategy makes the white space an infrequently encountered character in training. The visual archetypes enable the networks to generate spaces, whereas other models, which use one-hot vectors as textual content representations, generate noise instead of a space. This is shown in~\cref{fig:line_qualitative}, where we report an example of line-level generation. Moreover, we can notice that VATr++ renders punctuation marks and rare characters more faithfully than the VATr baseline, thanks to the input preparation and regularization strategies introduced.
\begin{table}[t]
    \footnotesize
    \centering
    \setlength{\tabcolsep}{.38em}
    \caption{Comparison between different HTG methods in terms of FID, KID and HWD score on the line-level variant of the IAM dataset. The values of the KID are multiplied by $10^2$.}
    \label{tab:linelevel_comp}
    \resizebox{.55\linewidth}{!}{
    \begin{tabular}{l c ccc}
    \toprule
                             && \textbf{FID}    & \textbf{KID}   & \textbf{HWD}  \\
    \midrule
    \textbf{HiGAN}           && 28.80           & ~1.94          & 1.19          \\
    \textbf{HiGAN+}          && 46.24           & ~3.63          & 1.59          \\
    \textbf{HWT}             && 27.50           & ~1.91          & 1.04          \\
    \textbf{VATr}            && \textbf{22.66}  & \textbf{~1.12} & \textbf{0.97} \\
    \textbf{VATr++}          && 24.51           & ~1.23          & 1.28          \\
    \bottomrule
    \end{tabular}}
\end{table}
\subsection{Generalization Capabilities} \label{subsec:generalization}
One of the main characteristics that would make an HTG model useful and applicable in practice is its ability to generalize to new handwritings and words. In fact, despite the training IAM dataset being large and diverse, it cannot capture all the existing words and handwriting styles. Additionally, the character distribution of the training data in the IAM dataset is very unbalanced~\cite{pippi2023handwritten}. Thus, there are a lot of characters that the HTG model is rarely exposed to during training. In this section, we test our model's robustness to data that are rarely seen during training or that are completely unseen in terms of handwriting style, textual content, or both. We compare our model's performance to that of HWT~\cite{bhunia2021handwriting} and VATr~\cite{pippi2023handwritten}, as those are the most comparable alternatives for being few-shot, transformer-based approaches.

\tit{Generalization to Long-Tail Characters}
To quantitatively measure how well our approach can generate rare characters, we use a State-of-the-Art HTR model to read the generated words and verify whether these can be read correctly, \ie~all the contained characters are rendered properly regardless of their frequency in the HTG training set. In particular, we exploit the TrOCR model~\cite{li2021trocr}, which has been trained on images of both typewritten and handwritten text in multiple languages, and measure its CER on the HTG-generated images. A low CER means that the model can produce readable text. In this experiment, first, we task the compared HTG approaches to generate \num{10000} words from the IAM corpus to evaluate the readability in a relatively easy setting (referred to as \textbf{IAM\textsubscript{all}} in the following). Second, \num{10000} words from the IAM corpus containing at least one long-tail character are generated (we refer to this set as \textbf{IAM\textsubscript{long-tail}} in the following). For these two sets, we also compute the CER on the corresponding real images to give an idea of the expected performance of TrOCR on these words. Furthermore, we generate \num{10000} words containing a lot of long-tail characters randomly selected from an external English corpus (this set is referred to as \textbf{Balanced} in the following). Finally, \num{10000} words containing strings of uniformly sampled characters are generated (we call this set \textbf{Gibberish}). Note that these latter sets do not have real counterparts.

\begin{figure}[t]
    \centering
    \includegraphics[width=\columnwidth]{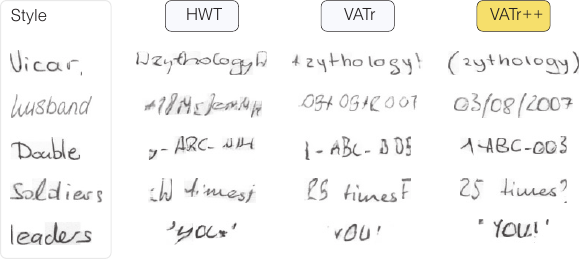}
    \caption{Example words containing rare characters rendered by using different HTG methods.}
    \label{fig:rare_qualitatives}
    \vspace{-.5em}
\end{figure}

The results of this experiment are reported in~\cref{tab:cer}. When generating common words from the IAM dataset (IAM\textsubscript{all}), VATr++ significantly outperforms both VATr and HWT, leading to a CER of only \num{0.79}. The same applies when the models are tasked to generate words with long-tail characters (IAM\textsubscript{long-tail}). It is also worth noting that, in both scenarios, the HTR model achieves lower CER on the words generated by VATr++ than on the real words. This is likely explained by the fact that the HTG models are explicitly optimized to generate words that are readable for an HTR model, although different from the model used for this test, whereas some real authors might have difficult-to-read handwritings. When using the HTG models to generate English words with a more uniform distribution of rare characters (the Balanced set), we also notice that VATr++ outperforms both HWT and VATr by a large margin. The same holds for the uniformly generated gibberish (the Gibberish set). Nonetheless, the CER values for this last set are much higher than in the others. The possible reason is twofold. First, even though VATr++ has been trained to generate rare characters better, it still struggles to generate strings that are very dissimilar to actual words, \ie~in which characters are combined in an unusual way, resulting in unusual ligatures. Second, HTR models struggle to read random text since they use an internal representation of language to improve their predictions. This applies also to our selected TrOCR model, although it has been trained on multiple languages. Nonetheless, from these results, we can conclude that VATr++ is significantly better at generating rare characters compared to previous approaches. This is also exemplified in~\cref{fig:rare_qualitatives}, which shows some words with rare characters generated by the considered HTG models.
\begin{table}[t]
    \footnotesize
    \centering
    \caption{CER value obtained by the State-of-the-Art TrOCR HTR model on images generated by different HTG models.}
    \label{tab:cer}
    \setlength{\tabcolsep}{.38em}
    \resizebox{.9\linewidth}{!}{
    \begin{tabular}{l c cccc}
    \toprule
                    && \textbf{IAM\textsubscript{all}}   & \textbf{IAM\textsubscript{long-tail}} & \textbf{Balanced}    & \textbf{Gibberish}   \\
    \midrule
    \textbf{Real}   && 4.13                              & 6.28                                  & -                    & -                    \\
    \midrule
    \textbf{HWT}    && 1.97                              & 6.98                                  & 16.68                & 82.52                \\
    \textbf{VATr}   && 1.84                              & 5.73                                  & 13.07                & 70.75                \\
    \textbf{VATr++} && \textbf{0.79}                     & \textbf{2.34}                         & \textbf{6.54}        & \textbf{63.89}       \\
    \bottomrule
    \end{tabular}}
\end{table}

\tit{Generalization to Unseen Vocabulary and Styles}
To assess the generalization capability in terms of unseen words and unseen writers' style, we consider the IV-S, IV-U, OOV-S, and OOV-U scenarios of the IAM dataset, as described in~\cref{sec:evaluation}. In this experiment, we compute the FID score on the images of each author separately and then average over all authors. Please refer to Appendix~\ref{app:results} for the KID and HWD metrics.

The results are reported in~\cref{tab:vocabulary} and indicate that VATr++ outperforms both VATr and HWT in the generation of images in the style of both seen and unseen writers and containing both in-vocabulary and out-of-vocabulary words. It is worth noting that all the considered models perform comparably when generating images with seen and unseen styles. This indicates good generalization capabilities towards the authors in the test set of IAM. Moreover, since the FID score is computed on a per-writer basis in this experiment, the lower value for VATr++ indicates that the model is also better at capturing the unique styles of each writer. Finally, we notice that a more visible difference can be appreciated between the IV and OOV settings, which remarks the impact of the content in the generated images (\ie~the characters in the word and their ligatures) on the score. 
\begin{table}[t]
    \footnotesize
    \centering
    \setlength{\tabcolsep}{.38em}
    \caption{FID scores obtained by testing our approach on unseen words and unseen writers.}
    \label{tab:vocabulary}
    \resizebox{.7\linewidth}{!}{
    \begin{tabular}{l c cccc}
    \toprule
                    && \textbf{IV-S}  & \textbf{IV-U}    & \textbf{OOV-S}    &   \textbf{OOV-U}  \\
    \midrule
    \textbf{HWT}    && 83.83          & 83.37            & 86.47             &   86.24           \\
    \textbf{VATr}   && 77.20          & 77.54            & 79.17             &   79.18           \\
    \textbf{VATr++} && \textbf{73.17} & \textbf{72.79}   & \textbf{74.51}    &   \textbf{74.27}  \\
    \bottomrule
    \end{tabular}}
\end{table}

\tit{Generalization to Other Datasets}
Finally, we evaluate the model performance when tested on datasets that are different from the training one. For this experiment, we consider the word-level CVL~\cite{kleber2013cvl} dataset and the line-level version of the RIMES~\cite{augustin2006rimes} dataset. Note that both datasets are multi-author, CVL contains words in English and German, and RIMES is in French.  For the CVL dataset, we define three settings: In-Vocabulary (\textbf{CVL-IV}), which considers words from CVL that also appear in the IAM training set; Out-of-Vocabulary (\textbf{CVL-OOV}), which considers words from CVL that do not appear in the IAM training set but are made of characters also present in the IAM charset; and Test (\textbf{CVL-Test}), which entails generating a replica of the CVL test set, including words with out-of-charset characters. For RIMES, we consider a setting in which the HTG model has to generate a replica of its test set (\textbf{RIMES-Test}). For this experiment we report FID values, KID and HWD values can be found in Appendix~\ref{app:results}.

The results for this experiment are shown in~\cref{tab:other_datasets}. We notice that the models exploiting Visual Archetypes outperform HWT, which is based on one-hot vectors, in all test sets. Moreover, it is worth noting that the FID values computed for the CVL settings are close to those obtained on the IAM dataset. This shows that our model generalizes well to a dataset with unseen styles and some unseen characters. On the other hand, the FID value on RIMES-Test is higher. We argue this is because the domain gap between RIMES and IAM is larger than that between CVL and IAM since RIMES contains only words in French while IAM is in English. 
\begin{table}[t]
    \footnotesize
    \centering
    \setlength{\tabcolsep}{.38em}
    \caption{FID scores obtained by testing on datasets that are different from the training one.}
    \label{tab:other_datasets}
    \resizebox{.9\linewidth}{!}{
    \begin{tabular}{l c cccc}
    \toprule
                    &&   \textbf{CVL-IV} & \textbf{CVL-OOV}  &   \textbf{CVL-Test}   &  \textbf{RIMES-Test}  \\
    \midrule
    \textbf{HWT}    &&   19.02           &   17.74           &   17.09               &   52.00               \\
    \textbf{VATr}   &&   \textbf{13.35}  &   14.49           &   \textbf{12.25}      &   \textbf{44.90}      \\
    \textbf{VATr++} &&   13.66           &   \textbf{13.52}  &   12.28               &   52.60               \\
    \bottomrule
    \end{tabular}}
\end{table}

\subsection{Ablation Analysis}
In this section, we analyze the impact of each of the strategies proposed to further boost the generation of long-tail characters. To this end, we train different versions of VATr++, each lacking one of the proposed strategies. We measure the model's ability to generate realistic images by measuring the FID for the IAM Test scenario described in~\cref{sec:evaluation}. To expose contributions towards rare character generation, we also measure the FID for the words from the IAM Test scenario that contain at least one long-tail character. For this scenario, we do not consider images of single punctuation marks. These images would contain long-tail characters that in IAM are scaled up to be as high as the other words and thus are disproportionate. Since our models are trained to generate punctuation marks that are proportional to other text, including singleton punctuation marks in the computation of the FID would result in a misleading value of the score. Moreover, note that long-tail characters often occupy a small portion of the word images and are located in the middle or at the end. For these reasons, their contribution to the final value of the FID, computed on the initial square crop of the image, is limited. Therefore, we also measure the CER of the State-of-the-Art TrOCR~\cite{li2021trocr} on words generated by each variant of VATr++. In particular, we generate \num{10000} words sampled from a corpus of English words containing multiple IAM long-tail characters in random styles sampled from the IAM test set, as done for the \textbf{Balanced} setting described in~\cref{subsec:generalization}. 

\begin{figure}[t]
    \centering
    \includegraphics[width=\columnwidth]{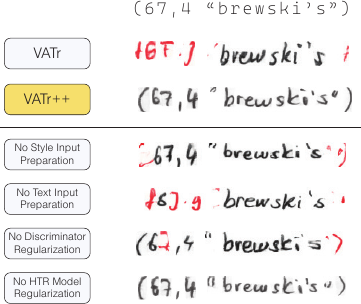}
    \caption{Qualitative ablation analysis of the main features of VATr++. Artifacts are colored in red.}
    \label{fig:ablation}
    \vspace{-.5em}
\end{figure}

Quantitative and qualitative results are reported in~\cref{tab:vatrpp_abl} and~\cref{fig:ablation}, respectively. Overall, VATr++ achieves lower FID and CER compared to VATr, which indicates its superior capability to reproduce realistic-looking handwriting faithfully and to generate images with readable text. From the ablation, we observe that when training the model without the style input preparation, the CER and the FID on words with long-tail characters increase. Arguably, this is because the presence of style images containing single punctuation marks hinders the network from learning what certain characters look like. This is also observed in the qualitatives, where the model struggles to generate brackets. Note that this variant achieves the best FID on the IAM Test setting, as it better reflects the original IAM dataset for not performing the style input preparation step. When not performing text input preparation in training, we observe a higher value for the CER. This demonstrates that the proposed text augmentation strategy helps train the model to generate all characters accurately. This is also reflected in the qualitatives, from which we observe that this variant of the model cannot generate most of the numbers and punctuation marks. As for the regularization strategies proposed, from the ablation, it emerges that the variant trained without discriminator regularization has the highest FID, highlighting the importance of this step when using text augmentation. Finally, we observe that not regularizing the HTR networks also leads to increased FID, especially on words with long-tail characters. This variant also achieves lower CER, indicating a tendency to overfit to IAM styles.
\begin{table}[t]
    \footnotesize
    \centering
    \setlength{\tabcolsep}{.58em}
    \caption{Quantitative ablation analysis of the main features of VATr++. }
    \label{tab:vatrpp_abl}
    \resizebox{\linewidth}{!}{
    \begin{tabular}{l c ccc}
    \toprule
                 & \textbf{FID\textsubscript{all}}    & \textbf{FID\textsubscript{long-tail}}  &   \textbf{CER}    \\
    \midrule
    \textbf{VATr++}                           &      16.29  &  22.24 & ~6.54 \\
    \midrule
    \textbf{no Style Input Preparation}       &      -0.44  &  +0.96 & +0.30 \\
    \textbf{no Text Input Preparation}        &      +0.06  &  +0.51 & +4.45 \\
    \textbf{no Discriminator Regularization}  &      +2.70  &  +2.77 & +0.16 \\
    \textbf{no HTR Model Regularization}      &      +0.65  &  +2.03 & -0.39 \\
    \bottomrule
    \end{tabular}}
\end{table}

\section{Conclusion}\label{sec:conclusion}
Current works in Styled HTG have mainly focused on better imitating the commonly adopted IAM dataset by achieving a lower FID score. In doing so, the faithful generation of rare characters, which are less represented in the IAM dataset, has been neglected. This has led to HTG systems that are less usable in practice. Additionally, there is no consistency in HTG literature on what part of the IAM dataset should be used for training and on how the FID value should exactly be computed. This has made it difficult to fairly compare different HTG methods, thus slowing down the research on this topic.

In this paper, we have extended our VATr architecture, a Styled HTG model that improves on long-tail character generation by using Visual Archetypes and that better captures handwriting style by exploiting a specific large-scale synthetic pretraining. In particular, we extended this method to VATr++ by introducing input preparation and training techniques that further improve the performance. Specifically, we have shown a major issue in how the IAM dataset is used for training HTG models and have proposed a specific style input preparation strategy to solve it. Moreover, we have proposed a text input preparation strategy to introduce more rare characters during training. Furthermore, we have proposed using random cropping to regularize the auxiliary discriminator network and an augmentation strategy to regularize the auxiliary HTR network. Additionally, we have introduced a standardized evaluation protocol that defines exactly what words to generate for evaluation and how to generate them, as well as which words to use as the reference for evaluation. A number of different scenarios are defined to capture different strengths and weaknesses of models. Using this framework, we have proved that our model can generate styled handwriting images across different scenarios and datasets, outperforming the competitors, especially when generating rare characters.

\section*{Acknowledgements}
This work was supported by the ``AI for Digital Humanities'' project (Pratica Sime n.2018.0390), funded by ``Fondazione di Modena'' and the PNRR project Italian Strengthening of Esfri RI Resilience (ITSERR) funded by the European Union – NextGenerationEU (CUP: B53C22001770006).

The internship of Bram Vanherle, during which this research was carried out, was supported by a grant by ``Fonds Wetenschappelijk Onderzoek - Vlaanderen (FWO)'' (File V421323N). Bram's PhD is supported by the Special Research Fund (BOF) of Hasselt University. The mandate ID is BOF20OWB24.

\bibliographystyle{IEEEtran}
\bibliography{main}

\newpage
\vspace{-30pt}
\begin{IEEEbiography}[{\includegraphics[width=1in,height=1.25in,clip,keepaspectratio]{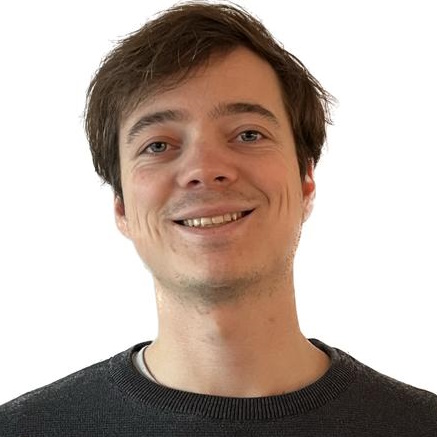}}]{Bram Vanherle} is a Ph.D. candidate from the University of Hasselt in Belgium. He is part of the Expertise center for Digital Media (EDM). He graduated from that same university in 2019, obtaining his M.Sc. in Computer Science. His current research topics range from handwritten text Generation to automated video directing, but his main focus is on synthetic training data for computer vision models.
\end{IEEEbiography}
\vspace{-1.cm}
\begin{IEEEbiography}[{\includegraphics[width=1in,height=1.25in,clip,keepaspectratio]{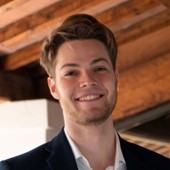}}]{Vittorio Pippi} received the M.Sc. degree in Computer Engineering cum laude from the University of Modena and Reggio Emilia, in 2020. He is currently a Ph.D. candidate of the National Ph.D. Program in Artificial Intelligence of the University of Pisa, with the University of Modena and Reggio Emilia as hosting university. His research interests include Document Analysis, Generative AI, and Digital Humanities.
\end{IEEEbiography}
\vspace{-1.cm}
\begin{IEEEbiography}[{\includegraphics[width=1in,height=1.25in,clip,keepaspectratio]{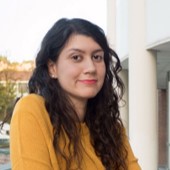}}]{Silvia Cascianelli} received the Ph.D. degree cum laude in Information and Industrial Engineering from the University of Perugia in 2019. She is an Assistant Professor with the University of Modena and Reggio Emilia. She was a Visitor Researcher at the Queen Mary University of London in 2018. Her research interests include Document Analysis, Digital Humanities, Generative AI, Vision and Language, and Embodied AI.
\end{IEEEbiography}
\vspace{-1.cm}
\begin{IEEEbiography}[{\includegraphics[width=1in,height=1.25in,clip,keepaspectratio]{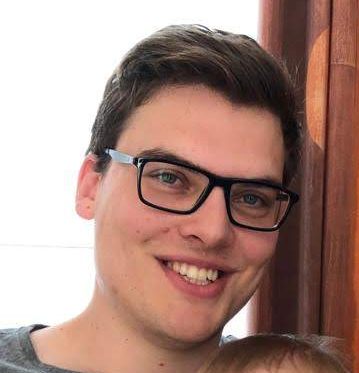}}]{Nick Michiels} received the Ph.D. degree in Computer Science from Hasselt university in 2016, where he is currently an assistant professor in the Faculty of Engineering Technology. He is affiliated with the Expertise Centre for Digital Media (EDM). He is one of the lead researchers in the Visual Computing group with expertise in the domains of computer vision, computer graphics, and image processing, including research on image-based rendering, XR, sim2real, and relighting.  
\end{IEEEbiography}
\vspace{-1.cm}
\begin{IEEEbiography}[{\includegraphics[width=1in,height=1.25in,clip,keepaspectratio]{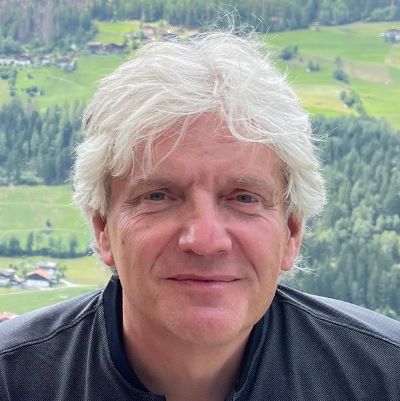}}]{Frank Van Reeth} received the Ph.D. degree in Computer Science from Hasselt university in 1993, where he is currently a Full Professor at the faculty of Sciences. He has supervised many PhD theses on diverse topics such as Computer Graphics, Animation, and Computer Vision. He is the Director of the Expertise center for Digital Media lab at that same university. His research interests are located in the domains of visual computing, robotics, computer graphics, and animation.
\end{IEEEbiography}
\vspace{-1.cm}
\begin{IEEEbiography}[{\includegraphics[width=1in,height=1.25in,clip,keepaspectratio]{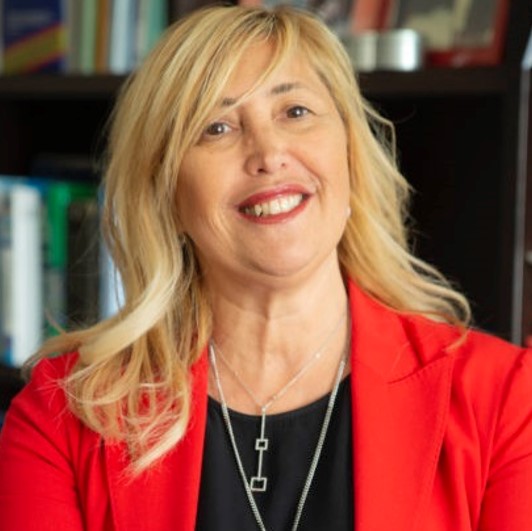}}]{Rita Cucchiara} received the Ph.D. in Computer Engineering from the University of Bologna in 1992. She is a Full Professor at the University of Modena and Reggio Emilia where she heads the AImageLab Laboratory. She serves and has served as Editor for international journals and Chair of international conferences on Computer Vision and Multimedia. Prof. Cucchiara is a Member of the ACM and the IEEE Computer Society, and a IAPR Fellow. She is Member of the Advisory Board of the Computer Vision Foundation and Director of the Modena ELLIS Unit and of the Artificial Intelligence Research and Innovation Center.
\end{IEEEbiography}

\vfill

\clearpage
\setcounter{page}{1}
\appendices
\section{Visual Archetypes Analysis} \label{app:archetypes}
\begin{figure}[h]
    \centering
    \includegraphics{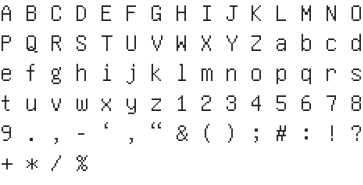}
    \caption{The Visual Archetypes used to encode character identity when training our model on the IAM dataset.}
    \label{fig:archetypes_examples}
\end{figure}

\begin{figure}[t]
    \centering
    \includegraphics[width=\columnwidth]{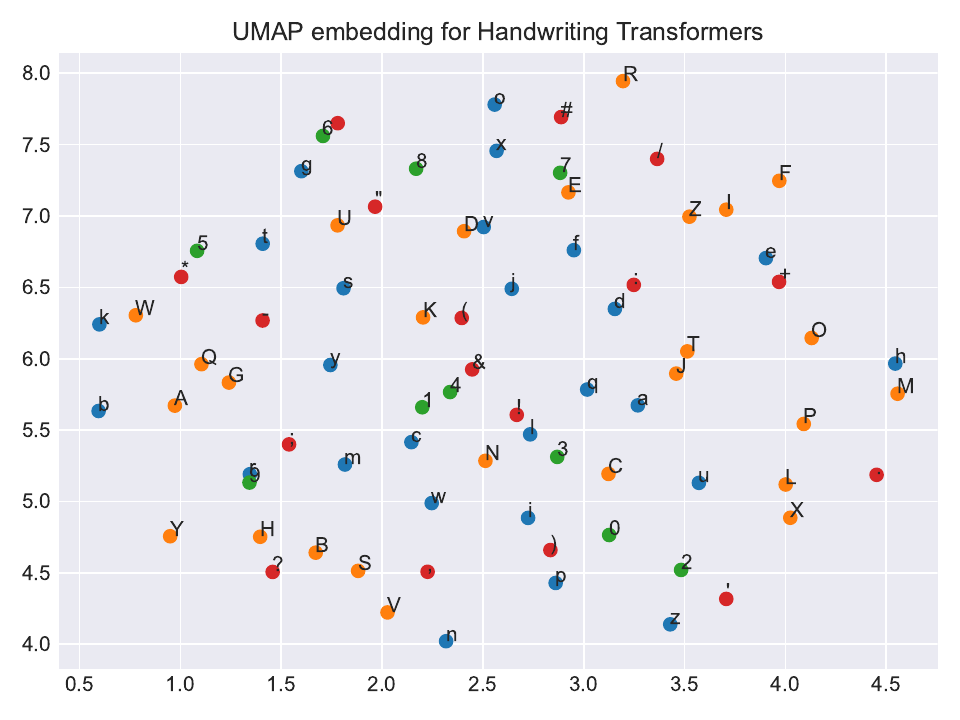}
    \includegraphics[width=\columnwidth]{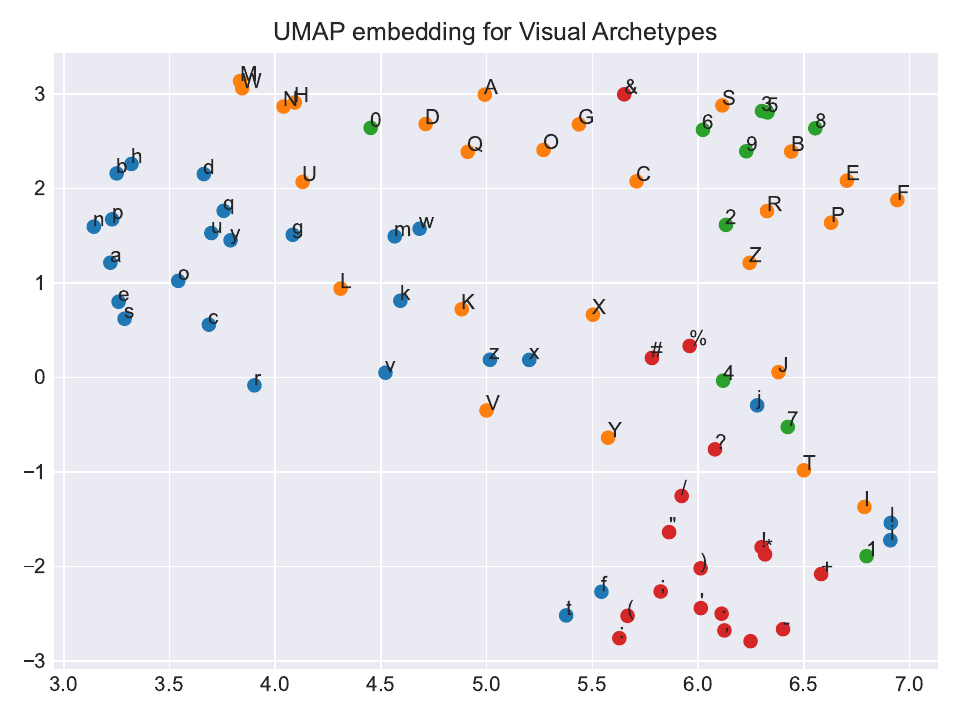}
    \caption{Comparison of the UMAP visualization of the character embeddings produced by both VATr (which uses Visual Archetypes) and HWT (which exploits one-hot encodings).}
    \label{fig:umap}
\end{figure}

In our seminal paper~\cite{pippi2023handwritten}, we introduced the Visual Archetypes as a better way to encode character identity for styled-HTG models. In this work, we still exploit Visual Archetypes. Note that previous works encoded the text input as independent one-hot vectors. This approach fails to exploit character similarity and limits the use of the model to generate characters seen at training time. Visual Archetypes consist of $16{\times}16$ binary images containing GNU Unifont characters. \cref{fig:archetypes_examples} shows all archetypes used while training the VATr++ model on the IAM dataset. To show that this technique helps the model to better associate similar characters, we obtain the UMAP visualization of the embeddings learned by the model for both VATr~\cite{pippi2023handwritten} and HWT~\cite{bhunia2021handwriting}, which uses a one-hot encoding. This is shown in~\cref{fig:umap}. It can be observed that the representations of similar characters via Visual Archetypes are better clustered than those exploiting one-hot vectors. Indeed, similar letters and punctuation marks are close to each other in the UMAP projection of Visual Archetypes-based embeddings, while for the embeddings based on the one-hot encoding, the disposition of different characters seems random.

\begin{figure}[h!]
    \centering
    \includegraphics[width=\columnwidth]{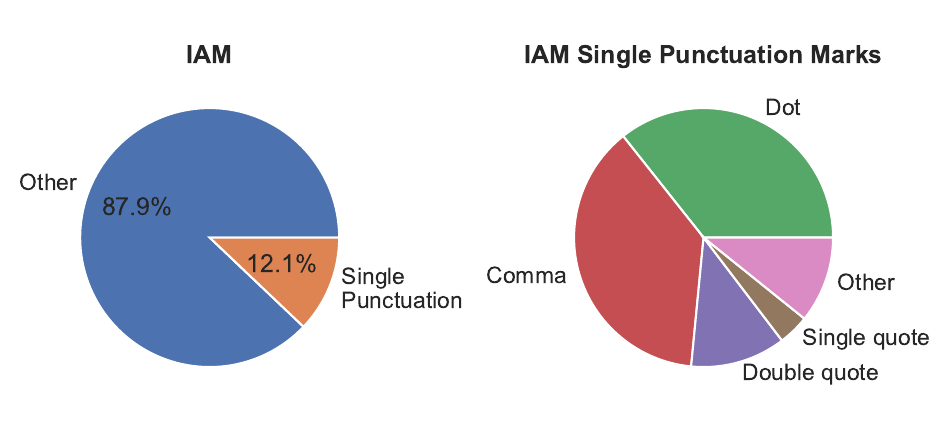}
    \caption{Portion of the IAM training dataset taken up by single punctuation marks (left) and distribution of the single punctuation marks in the IAM training set (right).}
    \label{fig:punctuation_distribution}
\end{figure}

\section{IAM Singleton Punctuation Marks} \label{app:single_punct}
In our paper, we identify the images containing singleton punctuation marks in the IAM dataset as one of the factors that make styled-HTG systems fail to render them, which are among the rare characters in the dataset. Note that the following punctuation marks occur in the IAM dataset: dot, comma, exclamation mark, question mark, opening bracket, closing bracket, double quote, single quote, colon, semi-colon, and dash. The IAM images containing them alone are a large portion of the IAM training set (12.1\%), with the dot (4.31\%) and comma (4.57\%) taking the biggest shares (see~\cref{fig:punctuation_distribution}, which reports a visual representation of this distribution). 

\begin{figure}[h]
    \centering
    \includegraphics[width=\columnwidth]{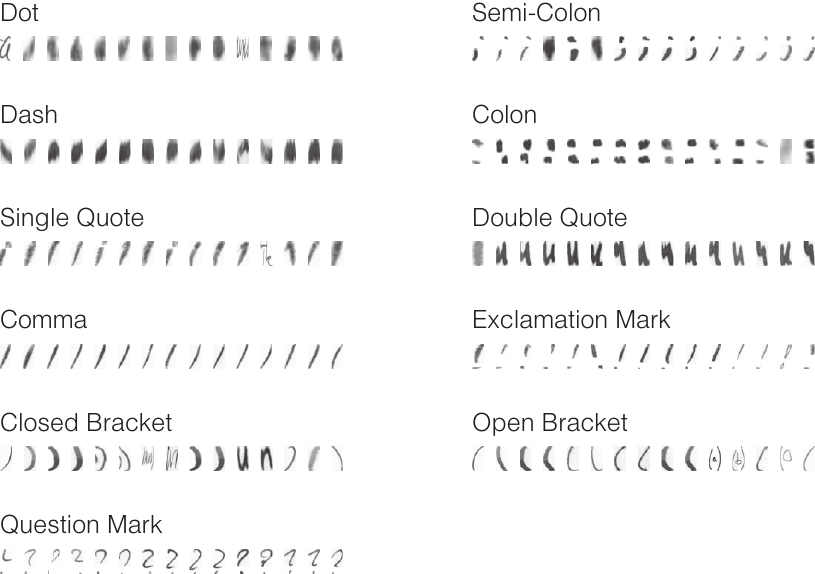}
    \caption{Examples of each of the single punctuation marks present in the IAM dataset.}
    \label{fig:punctuation_examples}
\end{figure}

Moreover, \cref{fig:punctuation_examples} shows 15 images containing each of the singleton punctuation marks in the IAM dataset. The images are reported after the transformations needed for them to be fed to the model, \ie~scaled to uniform height and width per character. It can be observed that these transformations can significantly alter small marks, such as dots, double quotes, and dashes, and make different punctuation marks look very similar (\eg~single quotes, commas, and brackets). Finally, note that the images reported have been selected purely based on their label. This explains some of the mistakes in the figure, such as the first example for the dot.

\section{Further Analysis of the Text Input Preparation} \label{app:input_prep}
\begin{figure}[h]
    \centering
    \includegraphics[width=\columnwidth]{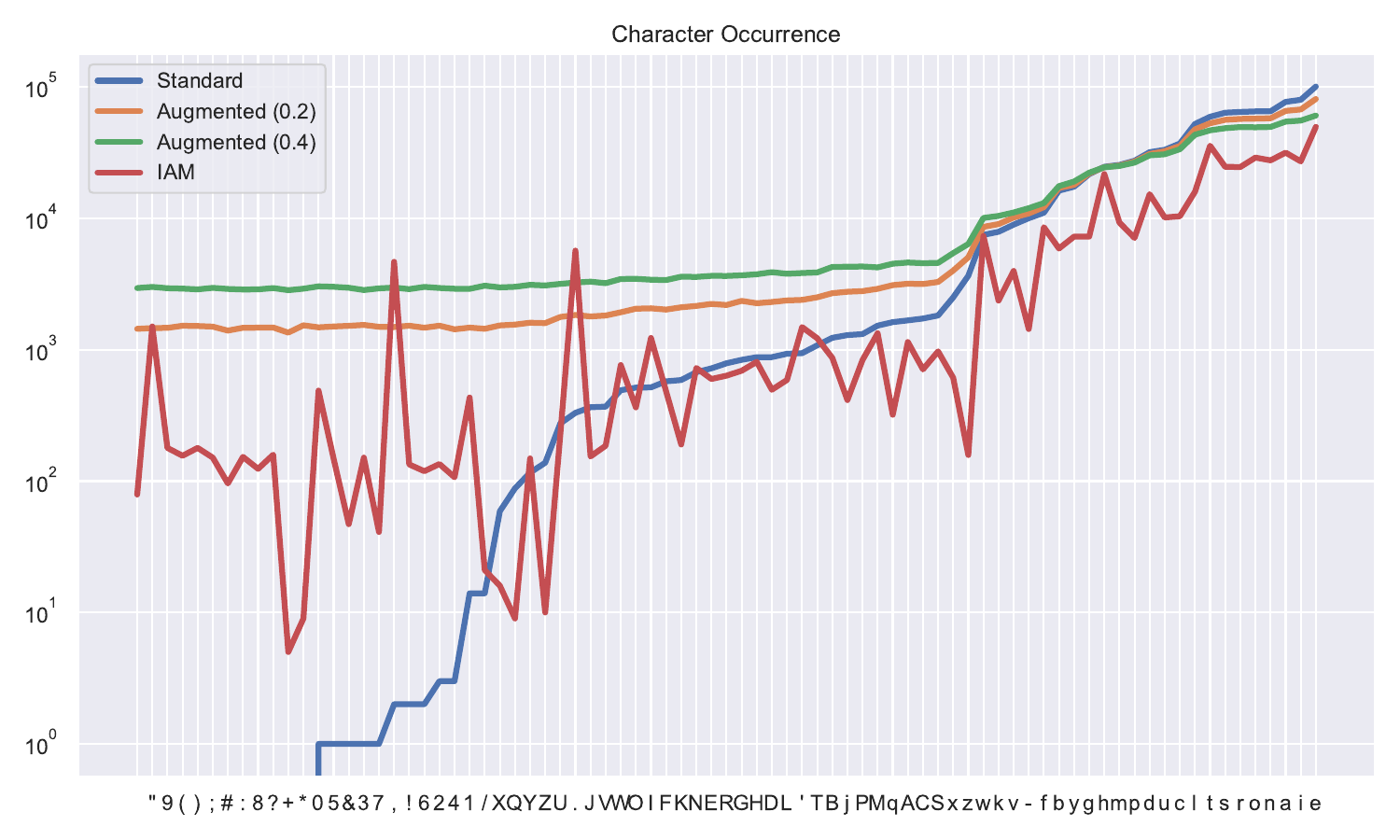}
    \caption{Distribution of characters for the standard corpus used to train styled-HTG methods, before and after applying our augmentation with two different values of strength $\alpha$ (in brackets). The character distribution of the IAM dataset is added for reference. The y-axis is in logarithmic scale.}
    \label{fig:character_distribution}
\end{figure}

The widely adopted approach to training a styled-HTG model entails asking it to generate words that are randomly selected from a predefined corpus of \num{466550} English words. The supervision signal for the model is computed over the styled rendering of these words. 
However, we observe that the character distribution of this predefined corpus is non-uniform, as can be observed from~\cref{fig:character_distribution}, which reports the character distribution of \num{100000} words sampled from the standard training corpus. Such skewed distribution can hinder the training of the HTG model by influencing the supervision signal. To mitigate this issue, we introduce a text input preparation strategy consisting in altering the characters of words in the training corpus proportionally to their original frequency. The effect of our input preparation strategy on the character distribution is also reported in~\cref{fig:character_distribution}. Moreover, \cref{tab:augmentation_examples} contains some examples of words being altered at different strengths $\alpha$.

\section{Further Analysis of the HTR Model Regularization} \label{app:htr_reg}
\begin{figure}
    \centering
    \includegraphics[width=\columnwidth]{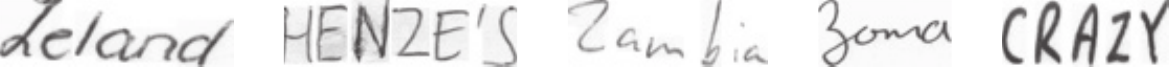}
    \caption{The only instances of words containing the capital letter Z in the IAM training set.}
    \label{fig:letter_z}
\end{figure}

The characters in the widely adopted IAM dataset appear in a long-tail distribution. As a result, there are only a few characters that are observed very often and in different styles, and some others that are observed rarely and possibly only in a few styles. This could cause the auxiliary HTR model that is trained alongside the generator to overfit on these limited instances of the rare characters. For example, \cref{fig:letter_z} shows all five instances of words containing the capital letter Z in the IAM training set. It can be observed that the styles are quite diverse. However, five samples are still not enough to train a model able to generalize. To combat this, we propose to regularize the HTR model by applying random transformations to its training input images. 

\begin{figure*}
    \centering
    \includegraphics[width=\textwidth]{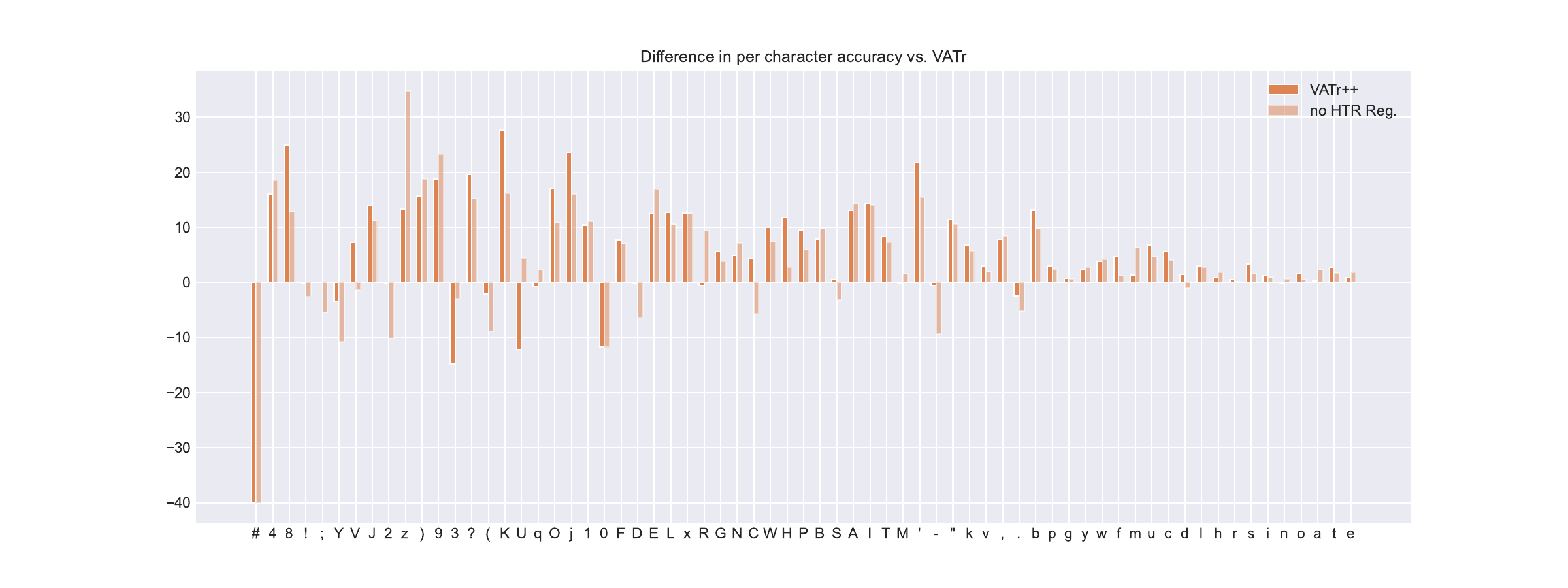}
    \caption{Difference in the character-wise accuracy of the HTR modules of VATr++ and VATr++ trained without HTR regularization, compared to the HTR model in VATr on the IAM test set.}
    \label{fig:htr_no_aug}
\end{figure*}

To test this assumption, we measure the character-wise accuracy of the HTR model of VATr++ over the words in the IAM dataset. In particular, we compute the character-wise accuracy by considering a character prediction as correct if it matches the ground truth character at the corresponding position. We do the same for a non-regularized version of the HTR model trained alongside VATr++. In~\cref{fig:htr_no_aug}, we report the difference in the character-wise accuracy between these models and the HTR network trained alongside the original VATr. We notice that for most characters, both the VATr++ HTR model and the non-regularized one outperform the VATr HTR model, with the former being generally better than the non-regularized version. From these results, we can conclude that the regularization is a helpful addition to the HTR module but that there are also other factors, such as the style input preparation, that lead VATr++ to outperform VATr.

\begin{table}[t]
    \footnotesize
    \centering
    \setlength{\tabcolsep}{.38em}
    \caption{Examples of words from the standard training set after text input preparation at different strengths $\alpha$.}
    \label{tab:augmentation_examples}
    \resizebox{.7\linewidth}{!}{
\begin{tabular}{c cc}
\toprule
\textbf{$\alpha$} & \textbf{Original} & \textbf{Altered} \\
\midrule
      \multirow{5}{*}{0.2} 
      & reforgive & refLrgive \\
	 & narcosynthesis & zaGc\&syntResis \\
	 & Husserl & Hussepl \\
	 & anda-assu & and6-as,u \\
	 & unretained & 6nretahned \\
\midrule
      \multirow{5}{*}{0.4} 
      & alchera &  cchera \\
	 & unromantically & u!r;man\#ica*ly \\
	 & loquency & loqucncy \\
	 & leapingly & l5*pingly \\
	 & brede & breue \\
\midrule
      \multirow{5}{*}{0.6} 
      & bulbiform & Tulbiform \\
	 & inco-ordinate & inco-oJ8iBat  \\
	 & emmarbling & emfar)ling \\
	 & Melinis & MGlinLs \\
	 & blabbermouth & b/abberm"uth \\
\midrule
      \multirow{5}{*}{0.8} 
      & wainscot-paneled & wKpnscoN-pxnPlJd \\
	 & yoe & YGx \\
	 & BD & BD \\
	 & chugalugs & chuOklug' \\
	 & NALGO & NALGO \\
\bottomrule
    \end{tabular}}
\end{table}

\section{Further Analysis on the Discriminator Regularization} \label{app:disc_reg}

When training, we noticed that the discriminator might exploit the characteristics of the training data, such as word lengths and the presence of certain characters, and be biased towards those characteristics instead of the perceptual realism of the input images. To analyze this aspect, we consider the discriminator trained alongside VATr and compute its score for \num{10000} words generated by that model.

\begin{figure}
    \centering
    \includegraphics[width=\columnwidth]{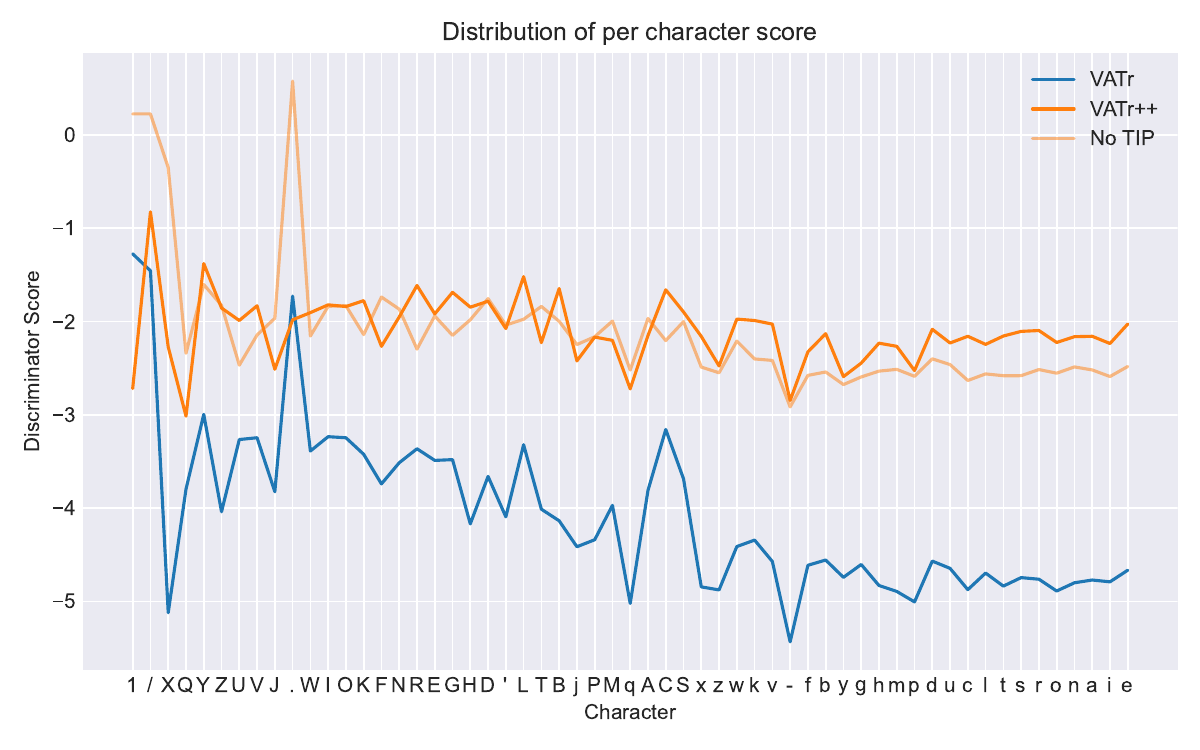}
    \caption{Average score of the discriminator for words containing each character. Computation is done over \num{10000} words generated by the model.}
    \label{fig:d_char_score}
\end{figure}

In particular, we estimate a per-character score by taking the average score of all word images containing that character. Ideally, we want this value to be roughly equal for all characters, as this would mean that the discriminator does not consider the textual content of its input images to determine whether they are real or generated. The distribution of the per-character scores is reported in~\cref{fig:d_char_score}. Note that the scores are ordered according to the frequency with which the characters appear in the standard, un-augmented training corpus. It can be observed that the score distribution is very uneven for the discriminator of VATr, which indicates a bias against or in favor of words containing certain characters. Moreover, we notice that this bias is correlated with the character frequency. 
On the other hand, the distribution of the score of the discriminator trained alongside VATr++ is more uniform. Arguably, this is thanks to the combination of the Discriminator Regularization strategy and the Text Input Preparation strategy adopted in VATr++. To isolate the impact of the discriminator regularization, we obtain the score distribution for a discriminator trained alongside a variant of VATr++ for which we do not perform Text Input Preparation. We notice that the distribution for this model is evened out and also shifted up compared to that of the VATr discriminator. From this, we can conclude that the proposed discriminator regularization strategy partly helps to overcome character bias, especially for some rare characters. Nonetheless, its main benefit is to make the prediction harder for the discriminator, which stabilizes the whole training process.

\begin{figure*}
    \centering
    \includegraphics[width=\columnwidth]{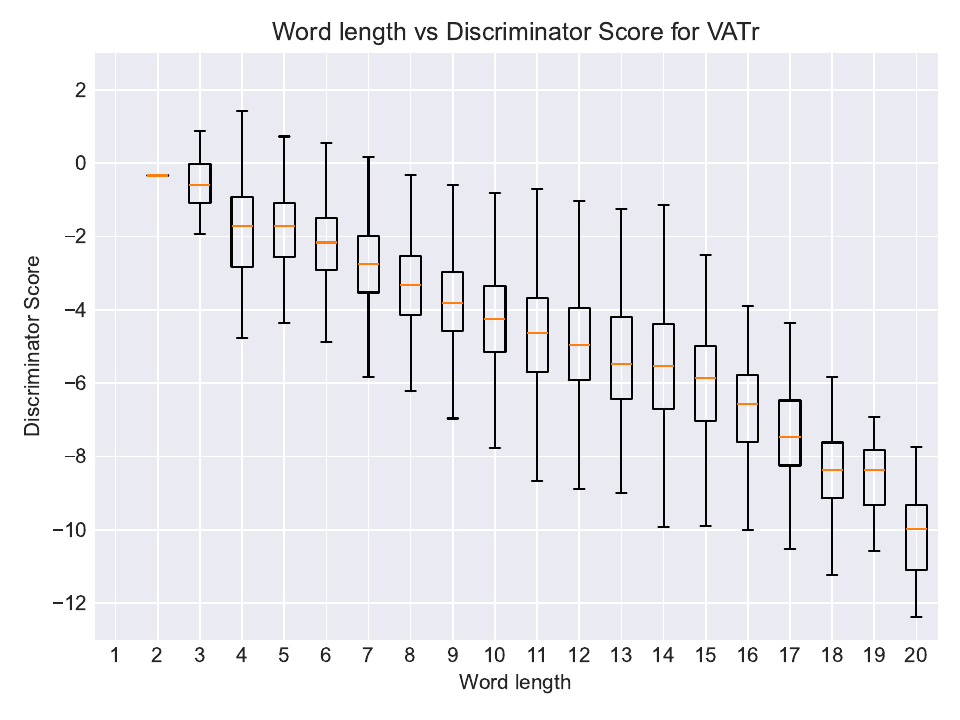}
    \includegraphics[width=\columnwidth]{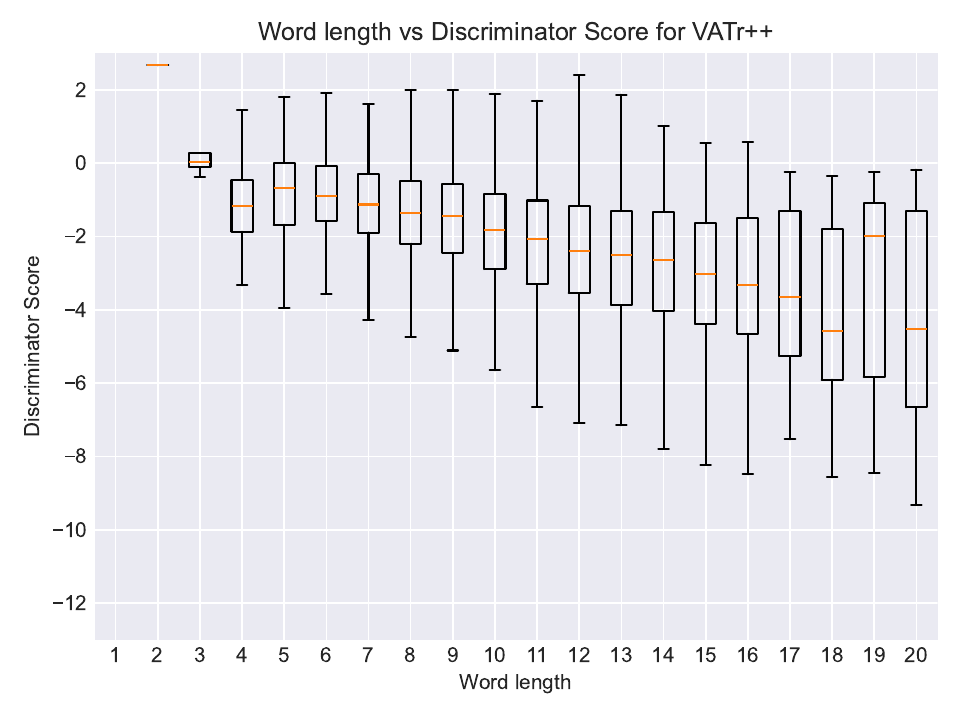}
    \caption{Average score of the discriminator for words of specific lengths generated by VATr and VATr++. Computation is done over \num{10000} words generated by the model.}
    \label{fig:d_length_scores}
\end{figure*}

Additionally, the discriminator shows a negative bias towards longer words, as shown in~\cref{fig:d_length_scores}. This is likely because longer words are not very common in the IAM dataset. Nonetheless, the regularized discriminator trained alongside VATr++ is much less impacted by this bias.

\begin{table*}[]
\footnotesize
\centering
\setlength{\tabcolsep}{.38em}
\caption{Results on the IV-S, IV-U, OOV-S and OOV-U scenarios of the IAM dataset. KID values are multiplied by $10^2$.}
\label{tab:oov}
\begin{tabular}{l c ccc c ccc c ccc c ccc}
\toprule
&& \multicolumn{3}{c}{\textbf{IV-S}} && \multicolumn{3}{c}{\textbf{IV-U}} && \multicolumn{3}{c}{\textbf{OOV-S}} && \multicolumn{3}{c}{\textbf{OOV-U}} \\
\cmidrule{3-5} \cmidrule{7-9} \cmidrule{11-13} \cmidrule{15-17}
&& \textbf{FID} & \textbf{KID} & \textbf{HWD} && \textbf{FID} & \textbf{KID} & \textbf{HWD} && \textbf{FID} & \textbf{KID} & \textbf{HWD} && \textbf{FID} & \textbf{KID} & \textbf{HWD} \\ 
\midrule
\textbf{HWT}    && 83.83 & 2.40 & 1.43 && 83.37 & 2.26 & 1.45 && 86,47 & 2.66 & 1.53 && 86.24 & 2.55 & 1.56 \\
\textbf{VATr}   && 77.20 & 1.82 & 1.29 && 77.54 & 1.74 & 1.33 && 79,17 & 1.91 & 1.40 && 79.18 & 1.81 & 1.45 \\
\textbf{VATr++} && \textbf{73.17} & \textbf{1.59} & \textbf{1.17} && \textbf{72.79} & \textbf{1.44} & \textbf{1.21} && \textbf{74,51} & \textbf{1.42} & \textbf{1.24} && \textbf{74.27} & \textbf{1.31} & \textbf{1.29} \\
\bottomrule
\end{tabular}
\end{table*}

\begin{table*}[]
\footnotesize
\centering
\setlength{\tabcolsep}{.38em}
\caption{Results on the CVL and RIMES datasets. KID values are multiplied by $10^2$.}
\label{tab:other_datasets}
\begin{tabular}{l c ccc c ccc c ccc c ccc}
\toprule
&& \multicolumn{3}{c}{\textbf{CVL-IV}} && \multicolumn{3}{c}{\textbf{CVL-OOV}} && \multicolumn{3}{c}{\textbf{CVL-Test}} && \multicolumn{3}{c}{\textbf{RIMES-Test}} \\
\cmidrule{3-5} \cmidrule{7-9} \cmidrule{11-13} \cmidrule{15-17}
&& \textbf{FID} & \textbf{KID} & \textbf{HWD} && \textbf{FID} & \textbf{KID} & \textbf{HWD} && \textbf{FID} & \textbf{KID} & \textbf{HWD} && \textbf{FID} & \textbf{KID} & \textbf{HWD} \\ 
\midrule
\textbf{HWT}    && 19.02 & 1.26 & 0.79 && 17.74 & 1.23 & 0.87 && 17.09 & 1.25 & 0.82 && 52.00 & 2.72 & 1.98 \\
\textbf{VATr}   && \textbf{13.35} & 0.81 & \textbf{0.71} && 14.49 & 0.89 & \textbf{0.69} && \textbf{12.25} & 0.80 & 0.72 && \textbf{44.90} & \textbf{1.94} & \textbf{1.48} \\
\textbf{VATr++} && 13.66 & \textbf{0.75} & 0.72 && \textbf{13.52} & \textbf{0.83} & 0.74 && 12.28 & \textbf{0.78} & \textbf{0.71} && 52.60 & 2.88 & 1.69 \\
\bottomrule
\end{tabular}
\end{table*}

\section{Additional Results} \label{app:results}

In the main paper, we only reported the FID value obtained when testing the different models for the IV-S, IV-U, OOV-S, and OOV-U scenarios of the IAM dataset. For completeness, in \cref{tab:oov}, we also report the HWD and KID scores for this experiment. Their values confirm the superiority of VATr++ with respect to the competitors.

Moreover, in the main paper, we also only reported the FID value for the analysis on the CVL and RIMES datasets. Thus, \cref{tab:other_datasets} also contains results expressed in terms of the HWD and KID scores. Also in this case, the effectiveness of the proposed strategies is confirmed by the good performance achievable, especially on English data.

\begin{figure*}[]
    \centering
    \includegraphics[width=\textwidth]{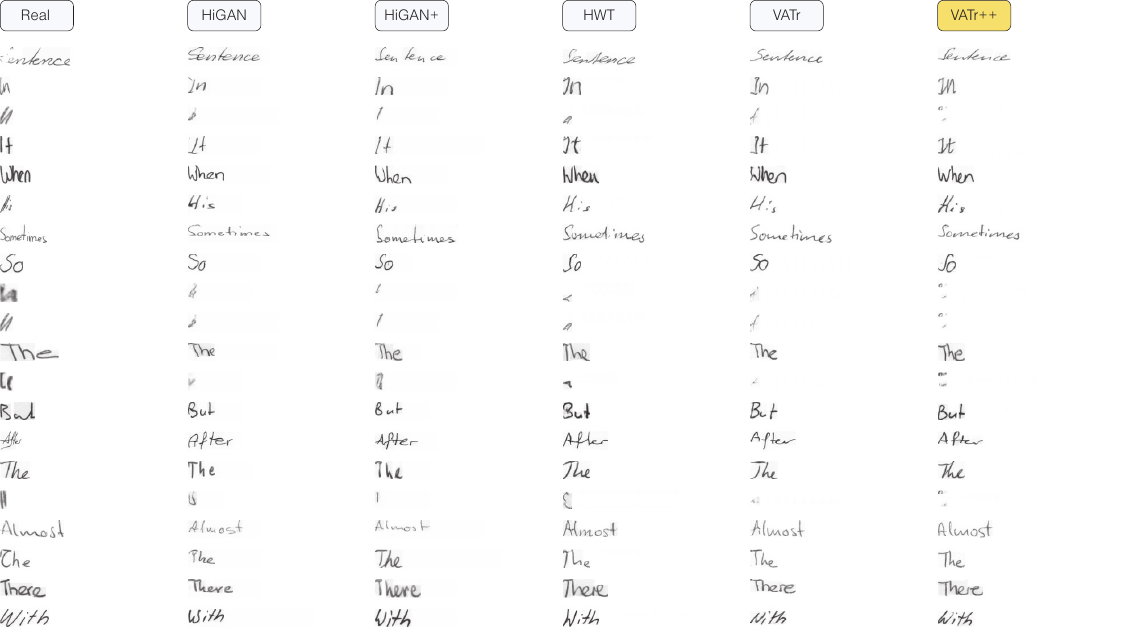}
    \caption{Qualitatives from the IAM dataset comparing VATr and VATr++ to other models such as HiGAN, HiGAN+ and HWT.}
    \label{fig:iam_qualitative}
\end{figure*}

\begin{figure*}
    \centering
    \includegraphics[width=\textwidth]{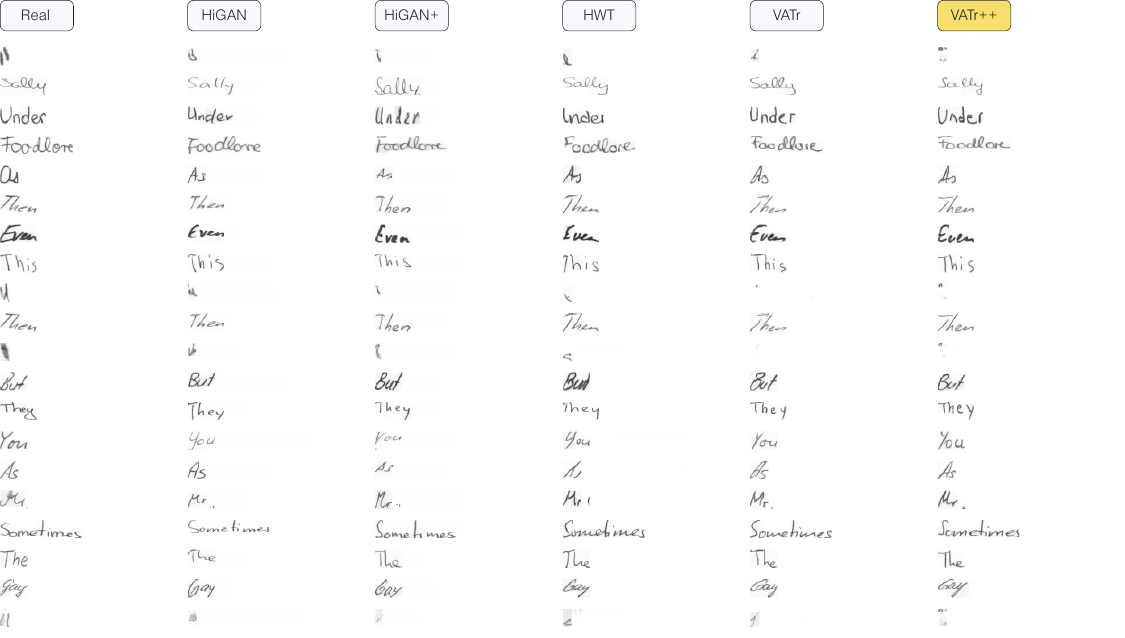}
    \caption{Qualitatives from the IAM-W16 dataset comparing VATr and VATr++ to other models such as HiGAN, HiGAN+ and HWT.}
    \label{fig:iam_w16_qualitative}
\end{figure*}

\begin{figure*}
    \centering
    \includegraphics[width=\textwidth]{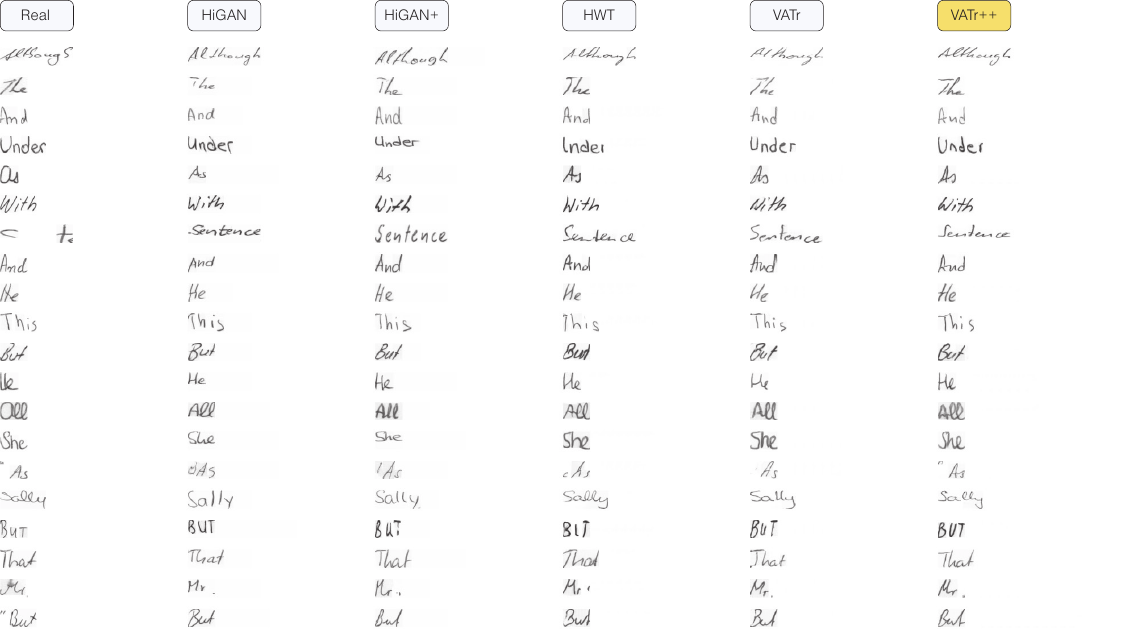}
    \caption{Qualitatives from the IAM-W16-ATTP dataset comparing VATr and VATr++ to other models such as HiGAN, HiGAN+ and HWT.}
    \label{fig:iam_w16_attp_qualitative}
\end{figure*}

\begin{figure*}
    \centering
    \includegraphics[width=\textwidth]{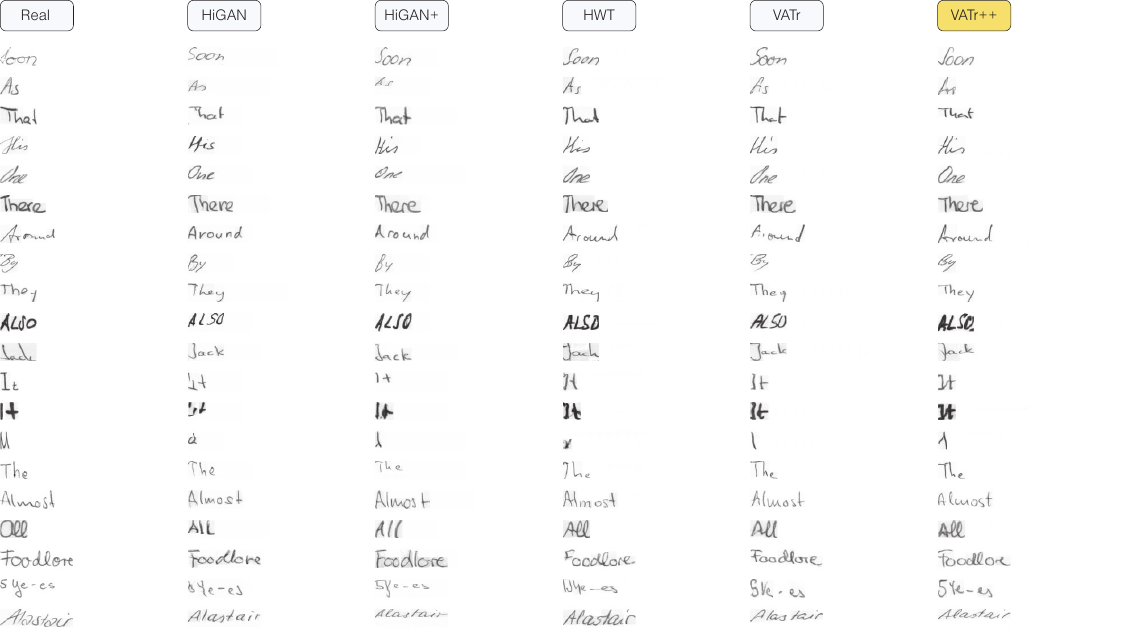}
    \caption{Qualitatives from the IAM-W16-NOP dataset comparing VATr and VATr++ to other models such as HiGAN, HiGAN+ and HWT.}
    \label{fig:iam_w16_nop_qualitative}
\end{figure*}

Finally, in the main paper, we compared our work to several other styled-HTG models: HiGAN, HiGAN+, HWT, and VATr. This comparison was done on a number of different versions of the IAM dataset: IAM-W, IAM-W16, IAM-WATTP, and IAM-WNOP. For a more complete comparison, \cref{fig:iam_qualitative,fig:iam_w16_qualitative,fig:iam_w16_attp_qualitative,fig:iam_w16_nop_qualitative} provide some additional qualitative results comparing our approach to the competitors on the different variants.

\end{document}